\documentclass{article} 
\usepackage{iclr2026_conference,times}

\usepackage{amsmath,amsfonts,bm}

\def\eqref#1{equation~\ref{#1}}

\def\1{\bm{1}}

\DeclareMathAlphabet{\mathsfit}{\encodingdefault}{\sfdefault}{m}{sl}
\SetMathAlphabet{\mathsfit}{bold}{\encodingdefault}{\sfdefault}{bx}{n}

\usepackage{hyperref}
\usepackage{url}
\usepackage{multirow}   
\usepackage{booktabs}   
\usepackage{graphicx}
\usepackage{subcaption}
\usepackage{wrapfig}

\title{DevelopmentODE: Structured Neural ODEs for Early Brain Development Dynamics Across a Decade }
\iclrfinalcopy
\author{
\begin{tabular}{@{}l@{}}
Kaiqiao Han$^{1}$, Haitao Chen$^{2}$, Bryan Quah$^{2}$,
Xiaoda Wang$^{1,3}$, Janelle Liu$^{2}$ \\
John H. Gilmore$^{4}$, Wei Gao$^{2}$, Yizhou Sun$^{1}$
\end{tabular}
\\[1ex]
$^{1}$University of California, Los Angeles \\
$^{2}$Cedars-Sinai Medical Center \\
$^{3}$Emory University \\
$^{4}$University of North Carolina at Chapel Hill
}

\begin{document}

\maketitle
\begin{abstract}

Understanding how individual brain development unfolds over childhood requires
modeling developmental trajectories from sparse longitudinal observations.
Long-term neurodevelopmental forecasting is not simply extending
short-term prediction to longer horizons as each child is typically observed at
only a few irregularly spaced visits and developmental dynamics vary
substantially with individuals and age. Generic continuous-time models naturally accommodate
irregular timing, but typically absorb these sources of variation into a single
flexible transition function, providing little structure for how
population-level progression, individual variability, and developmental age
should shape the dynamics.
We propose DevelopmentODE, a structured continuous-time framework that
organizes population and subject-specific variation within a shared
developmental geometry while allowing the governing dynamics to evolve with
developmental age. The model establishes this geometry around a developmental
canal representing the population-level trajectory, whose local direction
provides a reference for organizing subject-specific variation. Subject
deviation velocities are constrained relative to this developmental direction,
while a shared nonlinear deviation field captures individual developmental
motion without disrupting population-level progression.
Building on the same geometry, DevelopmentODE captures developmental
non-stationarity through ordered age-dependent deformations of the shared
vector field. These deformations progressively adapt a common underlying
dynamical structure as developmental age changes, while elapsed time determines
how long the resulting dynamics are integrated. The resulting formulation leverages broadly supported population-level
developmental structure to guide learning from sparse individual trajectories,
while allowing the governing dynamics to evolve smoothly across developmental
age.
We evaluate DevelopmentODE on longitudinal fMRI data by predicting future
functional connectivity observations of the same child from earlier
developmental observations. DevelopmentODE consistently outperforms competing baselines
across short- and long-horizon predictions. These results support the benefit
of explicitly structuring developmental dynamics for long-horizon
neurodevelopmental forecasting. Our code is publicly available.\footnote{\url{https://anonymous.4open.science/r/DevelopmentODE_anonymous-0CD7/}}

\end{abstract}

\section{Introduction}

Understanding how the human brain develops over time is a central question in neuroscience, particularly during childhood, when large-scale neural systems undergo substantial reorganization. Functional connectivity (FC), derived from functional magnetic resonance imaging (fMRI), provides a non-invasive characterization of coordinated activity across brain regions and has been associated with cognitive, behavioral, and neurodevelopmental outcomes~\citep{GRAYSON201715,chen2026functional,VANDENHEUVEL2010519}.
Most computational approaches to FC modeling, however, focus on cross-sectional data or relatively short temporal intervals. Such settings provide only a limited view of developmental dynamics, which unfold over years and may vary substantially across different stages of childhood. Long-horizon forecasting introduces a qualitatively different modeling regime: observations are sparse and irregular, developmental mechanisms change with age, and individual trajectories exhibit substantial variability despite sharing population-level developmental structure. Consequently, extending short-term predictive models to decade-long development is not simply a matter of extrapolating farther in time. It provides further insight that reflects the structure of the developmental process~\citep{COSIOGUIRADO2024101438,10.1093/cercor/bhad288,CHEN2021100976,chen2026functional}.

Longitudinal neuroimaging provides a unique window into individual
neurodevelopment, but learning developmental trajectories from sparse
measurements presents several key challenges.
(1) {Sparse and irregular observations.}
Longitudinal neuroimaging typically provides only a small number of visits per
subject, with heterogeneous intervals between measurements. Discrete sequence
models therefore observe sparse snapshots of an underlying continuous process
and are not naturally defined at arbitrary prediction times. Continuous-time
models provide a natural foundation for handling such observations, but do not
by themselves specify how different sources of developmental variation should
enter the dynamics
~\citep{10913768,VINCIBOOHER2026464,rubanova2019latentodesirregularlysampledtime}.
(2) {Population and individual variation are supported by different
amounts of evidence.}
Population-level developmental trends can be estimated from observations
across many subjects, whereas subject-specific deviations are supported by far
fewer repeated measurements~\citep{11356813,CHEN2021118079}. A single
unconstrained vector field must nevertheless account for both sources of
variation within the same flexible dynamics.
(3) {Developmental dynamics are non-stationary.}
Developmental age and elapsed time play distinct roles in longitudinal
evolution: when a transition occurs matters in addition to how long it lasts.
For example, a two-year transition from ages $1$ to $3$ need not follow the
same dynamics as one from ages $8$ to $10$
~\citep{10.1093/cercor/bhad288}. 

These properties motivate a continuous-time model with explicit structure for how
developmental variation is organized and evolves over age. We propose
{DevelopmentODE}, a framework that organizes population and
subject-specific dynamics within a shared developmental geometry while allowing
the governing dynamics to vary systematically with developmental age.
DevelopmentODE establishes a {population-shared developmental geometry}
around a developmental canal, which represents the population-level
trajectory and whose local direction provides a reference for organizing
subject-specific variation. Subject deviation velocities are modeled in the
complementary geometry, where a shared nonlinear deviation field captures
individual developmental motion while preserving population-level developmental
progression. This structure couples individual dynamics to broadly supported
population-level development while allowing longitudinal transitions to be
pooled across subjects.
Building on the same geometry, DevelopmentODE captures developmental
non-stationarity through {ordered age-dependent deformations of the
shared vector field}. These deformations smoothly adapt a common underlying
dynamical structure across developmental age without introducing independent
dynamics for each stage. Developmental age determines how the deformations are
combined, while elapsed time determines how long the resulting dynamics are
integrated. Smooth variation across ordered age ranges therefore yields
continuous age-dependent dynamics while preserving shared structure across
developmental stages.
DevelopmentODE thus provides a unified continuous-time formulation for sparse
and irregular longitudinal observations, in which individual trajectories are
organized relative to population-level progression and the governing dynamics
adapt continuously with developmental age.

We evaluate DevelopmentODE on longitudinal fMRI data for multi-year
within-subject functional connectivity forecasting. Given a child's functional
connectivity at one developmental age, the model predicts the same child's
connectivity at a future age. We consider both an all-forward setting and a
more restrictive adjacent-transition setting, where models are trained only on
consecutive observations but evaluated across all future intervals.
DevelopmentODE consistently outperforms baselines across short- and
long-horizon predictions, including direct predictors, sequential models,
Neural ODEs, and age-conditioned continuous-time models. Controlled ablations
further validate the contributions of the population-guided developmental
geometry and age-dependent dynamics.

Our contributions are threefold. (1) We formulate multi-year neurodevelopmental forecasting as a structured continuous-time problem that distinguishes population-level from subject-specific variation and developmental age from elapsed time. (2) We propose \textbf{DevelopmentODE}, which organizes subject-specific dynamics around a population-shared developmental geometry and captures developmental non-stationarity through ordered age-dependent modulation. (3) We evaluate DevelopmentODE on multi-year within-subject FC forecasting under all-forward and adjacent-transition settings, showing consistent improvements over direct, sequential, and continuous-time baselines, with ablations supporting the proposed geometry and age-dependent dynamics.

\section{Related Work}

\subsection{Infant and Child Brain Development}
\label{biob}
Brain development involves continuous but non-uniform reorganization of large-scale neural systems. Longitudinal neuroimaging studies have shown that functional connectivity (FC) follows heterogeneous developmental trajectories, with different brain systems maturing at different rates and over different developmental periods. Primary sensory and motor systems generally mature earlier, whereas higher-order association systems exhibit more prolonged development. Such observations are consistent with the notion of {developmental heterochrony}, in which the timing and rate of maturation vary across developmental stages. Consequently, equal elapsed time at different ages need not correspond to equivalent developmental change~\citep{CHEN2021100976,10.1093/cercor/bhad288, KEYTE201499}.
Development is also characterized by substantial inter-individual variability around shared population-level patterns. Such population-level regularity amid individual variability is consistent with the biological principle of developmental canalization, which describes the tendency of development to follow relatively stable pathways despite individual and environmental variation~\citep{LOISON20194,doi:10.1073/pnas.102303999}. 



\subsection{Neural ODEs for Dynamical Systems}

Neural Ordinary Differential Equations (Neural ODEs) model continuous-time
latent dynamics through
$
\frac{d\mathbf{z}_i(t)}{dt}
=
g_\theta(\mathbf{z}_i(t),t),
$
where the neural vector field $g_\theta$ specifies the instantaneous evolution
of subject $i$. Given an observed state $\mathbf{z}_i(t_0)$, numerical ODE
solvers integrate the dynamics to a target time $t_1$, enabling prediction at
arbitrary and irregularly sampled time points
~\citep{3327757.3327764,rubanova2019latentodesirregularlysampledtime}.
This makes Neural ODEs a natural foundation for longitudinal modeling.
Their flexibility, however, does not explicitly reflect the structure of
long-term development. Developmental age and elapsed time play distinct roles,
while population-shared maturation and subject-specific variation are supported
by substantially different amounts of evidence. Generic age- or
subject-conditioned Neural ODEs can absorb these signals arbitrarily into a
single flexible vector field. DevelopmentODE instead structures how they enter
the dynamics: population-level development defines a shared developmental
reference, subject-specific dynamics are modeled relative to this structure,
and age-dependent variation modulates the shared dynamics over development.
The goal is therefore to introduce a bio-informed inductive bias for organizing
developmental variation within continuous-time dynamics, rather than simply
increasing model expressiveness.
\section{Method}
\label{sec:method}

\begin{figure*}[!t]
    \centering
    \includegraphics[width=\textwidth, trim=0 0 0 0, clip]{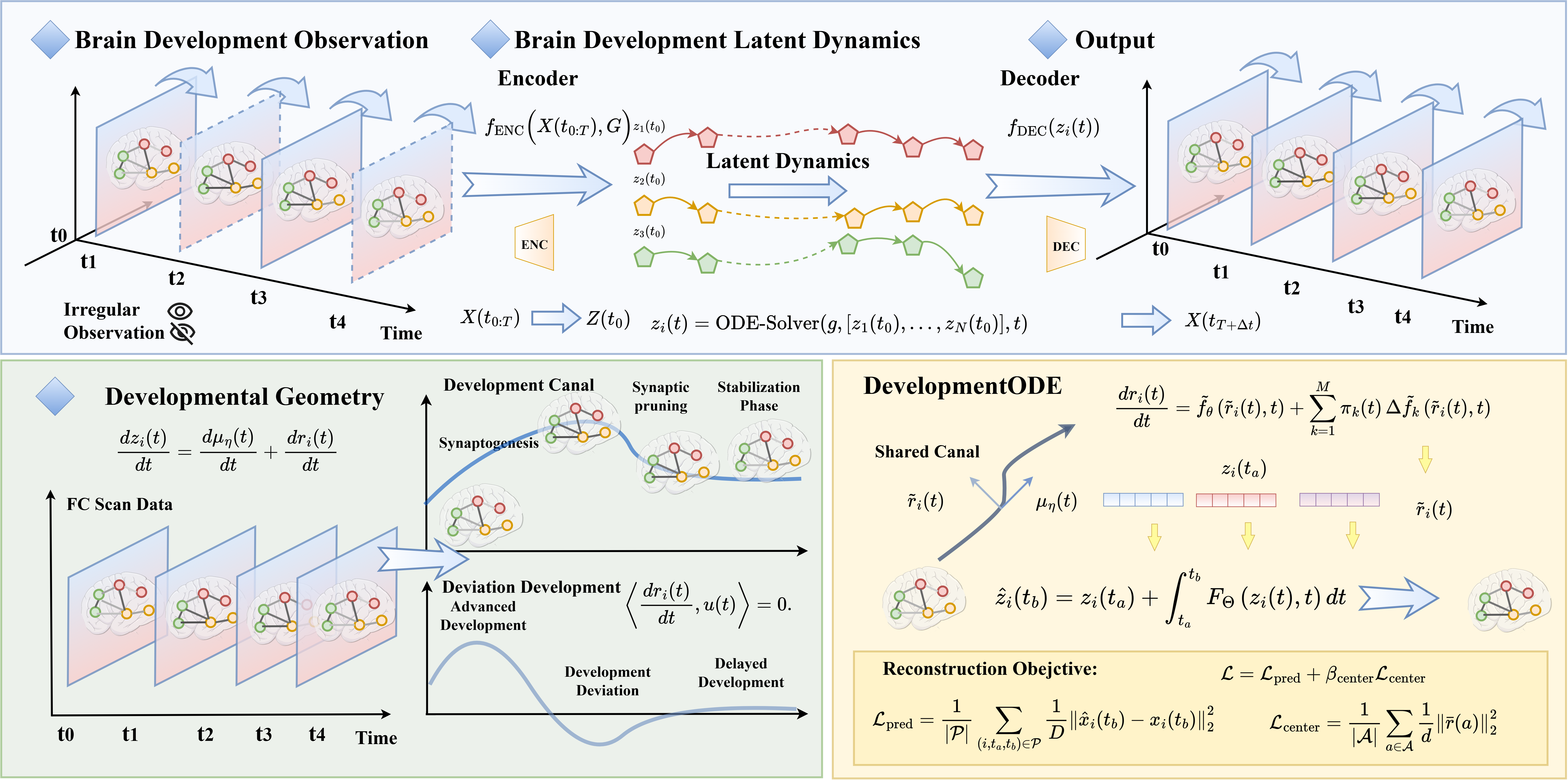}
\caption{
\textbf{Overview of DevelopmentODE.}
Sparse and irregular longitudinal FC observations are mapped into a shared
developmental geometry, where DevelopmentODE models continuous trajectories
and predicts future brain states at arbitrary target ages. Latent dynamics are
organized around a population developmental canal $\mu_\eta(t)$, with
subject-specific evolution modeled as deviation motion in the complementary
geometry (bottom left). Smoothly ordered age-dependent modulation adapts the
shared deviation field across development (bottom right). 
}
    \label{fig:intro}
\end{figure*}
We introduce DevelopmentODE, a structured continuous-time framework for learning non-stationary developmental dynamics from sparse longitudinal observations. The model organizes latent dynamics around a differentiable developmental canal that represents population-level progression and defines a local developmental geometry. Subject-specific motion is modeled relative to this shared trajectory by a nonlinear deviation field operating in the complementary space. To capture developmental non-stationarity, DevelopmentODE further applies smoothly ordered age-dependent modulation to the shared deviation field, yielding a continuous family of developmental dynamics parameterized by age while preserving shared structure across stages.
\subsection{Problem Formulation}
\label{sec:problem}

Consider a longitudinal cohort in which subject $i$ is observed at a sparse and
possibly irregular sequence of developmental ages,
\begin{equation}
\mathcal{X}_i
=
\left\{
\left(X_i(t_i^1),t_i^1\right),
\ldots,
\left(X_i(t_i^{n_i}),t_i^{n_i}\right)
\right\},
\end{equation}
where $X_i(t)\in\mathbb{R}^{K\times K}$ denotes the functional connectivity
(FC) matrix observed at developmental age $t$. Given a source observation
$X_i(t_a)$, our goal is to predict the same subject at a future age $t_b>t_a$,
i.e.,
$\hat X_i(t_b)=\mathcal{F}_{\Theta}(X_i(t_a),t_a,t_b)$.
Development is indexed by {absolute age}, rather than only by elapsed
time. Thus, two intervals of equal duration can correspond to different
developmental transformations when they occur at different ages. The desired
dynamics must therefore allow the governing vector field to vary with
developmental age.
Because FC observations are high-dimensional relative to the available
longitudinal supervision, we perform dynamical modeling in a low-dimensional
latent space. Let $x_i(t)\in\mathbb{R}^{D}$ denote the vectorized FC matrix.
A PCA basis $W\in\mathbb{R}^{d\times D}$ and mean $m\in\mathbb{R}^{D}$ are
estimated using training observations only:
$
z_i(t)
=
W\left(x_i(t)-m\right)$ for $
z_i(t)\in\mathbb{R}^{d}.
\label{eq:latent_state}
$
Predicted latent states are mapped back to FC space as
$\hat x_i(t)=W^\top \hat z_i(t)+m$.

\subsection{Structured Developmental Dynamics and Geometry}
\label{sec:canal}

Population-level developmental structure is supported by substantially more observations than individual longitudinal transitions. Rather than asking a single unconstrained field to explain both sources of variation, we explicitly model the population component and use it to define the geometry in which subject-specific dynamics evolve. This assigns shared developmental progression and individual variation to different dynamical degrees of freedom. We therefore represent
this component explicitly using a differentiable developmental canal
$\mu_\eta:t\mapsto\mathbb{R}^{d}$. Each latent dynamic is decomposed as
\begin{equation}
\frac{d z_i(t)}{dt}
=
\frac{d \mu_\eta(t)}{dt}
+
\frac{d r_i(t)}{dt},
\label{eq:canal_decomposition}
\end{equation}
where $\mu_\eta(t)$ represents the shared population trajectory and $r_i(t)$
captures subject-specific deviation from this trajectory.
We initialize the canal from population statistics in the training set. For
each observed developmental age $a\in\mathcal{A}$, let
$
\bar z(a)
=
\frac{1}{N_a}
\sum_{(i,t):t=a}
z_i(t),
$
where $N_a$ denotes the number of training observations at age $a$. We initialize population dynamics using
$
\mathcal{L}_{\mathrm{canal}}
=
\frac{1}{|\mathcal{A}|}
\sum_{a\in\mathcal{A}}
\left\|
\mu_\eta(a)-\bar z(a)
\right\|_2^2.
\label{eq:canal_initialization}
$
Weighting developmental ages equally prevents densely sampled ages from
dominating the fitted trajectory.
Since $\mu_\eta(t)$ is differentiable, its derivative
$\dot{\mu}_\eta(t)=d\mu_\eta(t)/dt$ defines a continuous population-level
direction and geometry of developmental progression.
We normalize this tangent
and remove the corresponding component from the subject deviation:
\begin{equation}
u(t)
=
\frac{\dot{\mu}_\eta(t)}
{\|\dot{\mu}_\eta(t)\|_2+\varepsilon},
\;
\tilde r_i(t)
=
r_i(t)
-
\left\langle
r_i(t),u(t)
\right\rangle u(t),
\;
\tilde f_\theta(r,t)
=
f_\theta(r)
-
\left\langle
f_\theta(r),u(t)
\right\rangle u(t).
\label{eq:deviation_geometry}
\end{equation}

Here, $\tilde r_i(t)$ denotes the deviation after removing its instantaneous
component along population developmental progression.
We apply the same geometric separation to the velocity generated, and the two operations have complementary roles. Using
$\tilde r_i(t)$ as the input prevents the individual field from conditioning
its dynamics on population-aligned deviation, while $\tilde f_\theta$ prevents
the field from generating additional velocity along the population
developmental direction.

The population canal and individual dynamics are therefore assigned to complementary local degrees of freedom: the canal governs shared developmental progression, while the projected field models residual subject-specific motion. This structured geometry prevents the higher-capacity individual field from freely re-explaining population-level development and provides the basis on which DevelopmentODE subsequently introduces age-dependent modulation of individual dynamics.
\subsection{Ordered Stage-Dependent Developmental Field}
\label{sec:deviation}

Individual developmental variation may evolve differently across age, making a single stationary deviation field overly restrictive. However, learning separate vector fields for different developmental stages would fragment the already sparse longitudinal supervision. We therefore model developmental non-stationarity by applying structured, stage-dependent deformations to a shared deviation field.
A general stage-adaptive formulation
can be written as
\begin{equation}
\frac{d r_i(t)}{dt}
=
\tilde f_\theta
\left(
\tilde r_i(t),t
\right)
+
\sum_{k=1}^{M}
\pi_k(t)\,
\Delta \tilde f_k
\left(
\tilde r_i(t),t
\right),
\label{eq:general_phase_field}
\end{equation}
where $\pi_k(t)$ specifies the contribution of developmental stage $k$.
Although learning an independent neural field for each developmental stage is expressive, it fragments the already sparse longitudinal supervision across
multiple stage-specific models. We therefore use the stage-dependent terms
$\Delta \tilde f_k$ as structured, capacity-controlled corrections to the
shared deviation field. In implementation, each $\Delta\tilde f_k$ combines a
stage-specific deformation $-\lambda_k\tilde r$ with a bounded
low-rank directional correction.
This design allows developmental stages to modulate subject-specific dynamics
while the higher-capacity field $\tilde f_\theta$ pools longitudinal
transitions across all developmental ages.
Developmental stages are naturally ordered along age. We therefore parameterize
stage membership using $M$ smoothly overlapping ordered stages rather than an
unconstrained mixture-of-experts router. Let
$b_1<b_2<\cdots<b_{M-1}$ denote the ordered transition centers and let $T>0$
control transition smoothness. We define
$q_k(t)=\sigma((t-b_k)/T)$ for $k=1,\ldots,M-1$, and construct the stage
weights as
\begin{equation}
\boldsymbol{\pi}(t)
=
\left[
1-q_1(t),\,
q_{1}(t)-q_{2}(t),\,
\ldots,\,
q_{M-2}(t)-q_{M-1}(t),\,
q_{M-1}(t)
\right].
\label{eq:ordered_phase_weights}
\end{equation}

By construction, $\pi_k(t)\geq0$ and
$\sum_{k=1}^{M}\pi_k(t)=1$, yielding smooth transitions between developmental
stages. The transition centers are fixed from the developmental age structure
of the training data and are never determined using evaluation subjects. We
therefore use the ordered stages as a parsimonious parameterization of
developmental non-stationarity rather than treating stage discovery itself as
the learning objective.
The resulting formulation gives developmental age a direct role in modulating
subject-specific dynamics around the population trajectory. The deviation dynamics are therefore non-stationary even
though the main individual field is shared across ages: developmental age
modulates a common dynamical system rather than selecting an entirely
independent vector field.

\subsection{DevelopmentODE}
\label{sec:full_ode}
Combining the population canal and stage-modulated deviation dynamics gives the full DevelopmentODE:
\begin{equation}
\frac{d z_i(t)}{dt}
=
\frac{d\mu_\eta(t)}{dt}
+
\tilde f_\theta
\left(
\tilde r_i(t),t
\right)
+
\sum_{k=1}^{M}
\pi_k(t)\,
\Delta \tilde f_k
\left(
\tilde r_i(t),t
\right),
\label{eq:development_ode}
\end{equation}

The three terms have distinct roles:
$\frac{d\mu_\eta(t)}{dt}$ captures population-level developmental progression,
$\tilde f_\theta\left(\tilde r_i(t),t\right)$ models deviation dynamics shared across subjects and ages, and
$\sum_{k=1}^{M}
\pi_k(t)\,
\Delta \tilde f_k
\left(
\tilde r_i(t),t
\right)$ captures stage-dependent deformation of individual developmental dynamics.
This decomposition matches model capacity to the available supervision.
The population canal exploits cross-subject observations across developmental
ages while the deviation field pools all available longitudinal transitions.

Given an observation at age $t_a$, its latent representation
$z_i(t_a)=W(x_i(t_a)-m)$ defines the initial condition. We integrate
Eq.~\ref{eq:development_ode} from $t_a$ to $t_b$:
\begin{equation}
\hat z_i(t_b)
=
z_i(t_a)
+
\int_{t_a}^{t_b}
F_\Theta
\left(
z_i(t),t
\right)
dt.
\label{eq:development_flow}
\end{equation}

Because $\mu_\eta(t)$  depends on absolute developmental
age, equal-length prediction intervals need not induce the same transformation.
DevelopmentODE therefore separates the developmental clock from the elapsed
integration interval.

\subsection{Learning Objective}
\label{sec:objective}

\paragraph{Prediction loss.} We construct forward source--target pairs from repeated observations of the
same subject,
$
\mathcal{P}
=
\left\{
(i,t_a,t_b):t_b>t_a
\right\}.
$
Our primary objective evaluates predictions after reconstruction into FC space:
\begin{equation}
\mathcal{L}_{\mathrm{pred}}
=
\frac{1}{|\mathcal{P}|}
\sum_{(i,t_a,t_b)\in\mathcal{P}}
\frac{1}{D}
\left\|
\hat x_i(t_b)-x_i(t_b)
\right\|_2^2.
\label{eq:prediction_loss}
\end{equation}

To preserve the interpretation of $\mu_\eta(t)$ as the population trajectory,
we encourage deviations to remain centered at each observed developmental age.
Define
$
\bar r(a)
=
\frac{1}{N_a}
\sum_{(i,t):t=a}
\left[
z_i(t)-\mu_\eta(a)
\right].
$
The corresponding centering objective is
\begin{equation}
\mathcal{L}_{\mathrm{center}}
=
\frac{1}{|\mathcal{A}|}
\sum_{a\in\mathcal{A}}
\frac{1}{d}
\left\|
\bar r(a)
\right\|_2^2.
\label{eq:center_loss}
\end{equation}

The complete training objective is
\begin{equation}
\mathcal{L}
=
\mathcal{L}_{\mathrm{pred}}
+
\beta_{\mathrm{center}}
\mathcal{L}_{\mathrm{center}}.
\label{eq:total_objective}
\end{equation}

\paragraph{Subject-balanced variable-interval training.}
\label{sec:training}
Subjects with more repeated observations yield more source--target pairs and
would be overrepresented under uniform pair sampling. We therefore first sample
an eligible subject uniformly and then sample one valid forward pair from that
subject.
Prediction intervals within the same minibatch may have different source and
target ages. To integrate these trajectories in parallel, we introduce a
normalized progress variable $u\in[0,1]$, with
$t_i(u)=t_{a,i}+u\Delta t_i$ and
$\Delta t_i=t_{b,i}-t_{a,i}$. Applying the chain rule gives
$
\frac{d z_i}{du}
=
\Delta t_i
F_\Theta
\left(
z_i(u),t_i(u)
\right),
$
which permits heterogeneous prediction intervals to be integrated jointly
while preserving their absolute-age dependence.

\section{Experiments}

We evaluate the proposed DevelopmentODE framework on a longitudinal fMRI dataset of children, aiming to predict future functional connectivity (FC) matrices across a decade. 
\subsection{Experiment Setup}

\paragraph{Dataset and Experimental Setup.}
We use the longitudinal resting-state fMRI dataset from the University of North Carolina Early Brain Development Study (EBDS), comprising 1,476 scans from 633 children acquired at approximately 3 weeks, 1, 2, 4, 6, 8, and 10 years. We retain $N=484$ subjects with at least two time points. FC matrices are computed from region-level time series using Pearson correlation and have size $278 \times 278$~\citep{chen2026functional}. We additionally evaluate DevelopmentODE on longitudinal rsfMRI data from the independently collected Baby Connectome Project (BCP), using the same modeling and evaluation pipeline~\citep{HOWELL2019891}. Additional dataset details are provided in Appendix~\ref{dataset}.
We use subject-level five-fold cross-validation, with approximately 65\%, 15\%, and 20\% of subjects assigned to training, validation, and test sets in each fold. We separately evaluate short-term ($\Delta t \leq 2$ years) and long-term ($\Delta t > 2$ years) forecasting. The full five-fold evaluation is repeated over three random seeds, with mean performance reported. Appendix~\ref{sec:implementation} provides the adjacent-transition evaluation and additional implementation details.

\paragraph{Baselines and Evaluation Metrics}
We compare DevelopmentODE with three groups of alternatives. First, \emph{direct predictors} include Direct MLP and Time-conditioned MLP ({Time-MLP}), which predict the target representation without explicitly modeling a continuous trajectory. Second, \emph{discrete sequence models} include RNN, GRU, LSTM, and Transformer~\citep{10.1162/neco.1997.9.8.1735,cho-etal-2014-learning,NIPS2017_3f5ee243}. Third, \emph{continuous-time models} include Generic Neural ODE ({Generic NODE}), Age-conditioned Neural ODE ({Age-NODE}), MoE Neural ODE ({MoE-NODE}), and Stage Neural ODE ({Stage-NODE})~\citep{11356813,10913768,rubanova2019latentodesirregularlysampledtime}. The names in parentheses correspond to the shortened model names used in Table~\ref{tab:main_results1}.
We evaluate prediction quality using four standard metrics: Mean Squared Error (MSE), which emphasizes larger prediction errors; Mean Absolute Error (MAE), which measures average element-wise error; Pearson Correlation Coefficient (PCC), which measures agreement between predicted and ground-truth FC values; and Coefficient of Determination ($R^2$), which measures the proportion of variance explained by the predictions.

\subsection{Main Results}

\begin{table*}[t!]
\caption{
Evaluation results on short-term, long-term, and all prediction pairs.
Metrics include MSE ($\downarrow$), MAE ($\downarrow$), $R^2$ ($\uparrow$),
and PCC ($\uparrow$).
The values following each metric denote the relative improvement (\%) over
the RNN baseline, where positive values indicate
better performance according to the direction of each metric.
The best results are in \textbf{bold}, and the second-best are
\underline{underlined}.
MSE, $R^2$, and PCC are multiplied by 100, and MAE by 10, for readability.
}
\label{tab:main_results1}

\centering
\scriptsize
\setlength{\tabcolsep}{2.8pt}
\renewcommand{\arraystretch}{1.08}

\newcommand{\chg}[1]{\scalebox{0.5}{#1}}

\resizebox{\textwidth}{!}{%
\begin{tabular}{@{}lcccccccccccc@{}}
\toprule
\textbf{Model}
& \multicolumn{4}{c}{\textbf{Short-term}}
& \multicolumn{4}{c}{\textbf{Long-term}}
& \multicolumn{4}{c}{\textbf{All}} \\
\cmidrule(lr){2-5}
\cmidrule(lr){6-9}
\cmidrule(lr){10-13}

& MSE $\downarrow$
& MAE $\downarrow$
& $R^2$ $\uparrow$
& PCC $\uparrow$
& MSE $\downarrow$
& MAE $\downarrow$
& $R^2$ $\uparrow$
& PCC $\uparrow$
& MSE $\downarrow$
& MAE $\downarrow$
& $R^2$ $\uparrow$
& PCC $\uparrow$ \\
\midrule

RNN
& 3.07
& 1.23
& 34.51
& 59.36
& 3.10
& 1.23
& 31.65
& 57.19
& 3.09
& 1.23
& 32.80
& 58.05 \\

GRU
& 3.09 \chg{-0.4\%}
& 1.23 \chg{0.0\%}
& 34.29 \chg{-0.6\%}
& 59.18 \chg{-0.3\%}
& 3.10 \chg{-0.1\%}
& 1.23 \chg{+0.1\%}
& 31.64 \chg{-0.1\%}
& 57.18 \chg{0.0\%}
& 3.10 \chg{-0.2\%}
& 1.23 \chg{+0.1\%}
& 32.69 \chg{-0.3\%}
& 57.97 \chg{-0.1\%} \\

LSTM
& 3.07 \chg{0.0\%}
& 1.22 \chg{+0.3\%}
& 34.59 \chg{+0.2\%}
& 59.40 \chg{+0.1\%}
& 3.09 \chg{+0.4\%}
& 1.22 \chg{+0.6\%}
& 31.96 \chg{+1.0\%}
& 57.43 \chg{+0.4\%}
& 3.08 \chg{+0.3\%}
& 1.22 \chg{+0.5\%}
& 33.01 \chg{+0.7\%}
& 58.22 \chg{+0.3\%} \\

Transformer
& 3.04 \chg{+1.3\%}
& \underline{1.22} \chg{+0.6\%}
& 35.32 \chg{+2.3\%}
& 60.02 \chg{+1.1\%}
& 3.02 \chg{+2.5\%}
& 1.21 \chg{+1.4\%}
& 33.42 \chg{+5.6\%}
& 58.47 \chg{+2.3\%}
& 3.03 \chg{+2.0\%}
& 1.21 \chg{+1.1\%}
& 34.17 \chg{+4.2\%}
& 59.09 \chg{+1.8\%} \\

Direct MLP
& 3.03 \chg{+1.5\%}
& 1.22 \chg{+0.2\%}
& 35.58 \chg{+3.1\%}
& 60.09 \chg{+1.2\%}
& 3.07 \chg{+1.1\%}
& 1.23 \chg{-0.1\%}
& 32.41 \chg{+2.4\%}
& 57.66 \chg{+0.8\%}
& 3.05 \chg{+1.2\%}
& 1.23 \chg{0.0\%}
& 33.68 \chg{+2.7\%}
& 58.63 \chg{+1.0\%} \\

Time-MLP
& \underline{3.02} \chg{+1.7\%}
& 1.22 \chg{+0.4\%}
& \underline{35.67} \chg{+3.4\%}
& \underline{60.16} \chg{+1.4\%}
& 3.05 \chg{+1.6\%}
& 1.22 \chg{+0.4\%}
& 32.80 \chg{+3.6\%}
& 57.98 \chg{+1.4\%}
& 3.04 \chg{+1.7\%}
& 1.22 \chg{+0.4\%}
& 33.95 \chg{+3.5\%}
& 58.85 \chg{+1.4\%} \\

Generic NODE
& 3.08 \chg{-0.3\%}
& 1.22 \chg{+0.2\%}
& 34.45 \chg{-0.2\%}
& 59.26 \chg{-0.2\%}
& 2.98 \chg{+3.8\%}
& 1.20 \chg{+2.0\%}
& 34.31 \chg{+8.4\%}
& 59.13 \chg{+3.4\%}
& 3.02 \chg{+2.2\%}
& 1.21 \chg{+1.3\%}
& 34.36 \chg{+4.8\%}
& 59.18 \chg{+1.9\%} \\

Age-NODE
& 3.08 \chg{-0.2\%}
& 1.22 \chg{+0.2\%}
& 34.49 \chg{-0.1\%}
& 59.30 \chg{-0.1\%}
& \underline{2.98} \chg{+3.8\%}
& 1.20 \chg{+1.9\%}
& \underline{34.32} \chg{+8.4\%}
& \underline{59.15} \chg{+3.4\%}
& 3.02 \chg{+2.2\%}
& 1.21 \chg{+1.2\%}
& 34.38 \chg{+4.8\%}
& 59.21 \chg{+2.0\%} \\

MoE-NODE
& 3.08 \chg{-0.1\%}
& 1.22 \chg{+0.2\%}
& 34.54 \chg{+0.1\%}
& 59.33 \chg{0.0\%}
& 2.98 \chg{+3.8\%}
& \underline{1.20} \chg{+2.0\%}
& 34.31 \chg{+8.4\%}
& 59.14 \chg{+3.4\%}
& \underline{3.02} \chg{+2.2\%}
& \underline{1.21} \chg{+1.3\%}
& \underline{34.39} \chg{+4.9\%}
& \underline{59.21} \chg{+2.0\%} \\

Stage-NODE
& 3.08 \chg{-0.1\%}
& 1.22 \chg{+0.2\%}
& 34.51 \chg{0.0\%}
& 59.45 \chg{+0.2\%}
& 3.01 \chg{+2.8\%}
& 1.21 \chg{+1.7\%}
& 33.63 \chg{+6.2\%}
& 58.68 \chg{+2.6\%}
& 3.04 \chg{+1.7\%}
& 1.21 \chg{+1.1\%}
& 33.98 \chg{+3.6\%}
& 58.99 \chg{+1.6\%} \\

\midrule

\textbf{DevelopmentODE}
& \textbf{2.93} \chg{+4.6\%}
& \textbf{1.21} \chg{+1.5\%}
& \textbf{37.51} \chg{+8.7\%}
& \textbf{61.56} \chg{+3.7\%}
& \textbf{2.94} \chg{+5.1\%}
& \textbf{1.20} \chg{+2.5\%}
& \textbf{35.26} \chg{+11.4\%}
& \textbf{59.82} \chg{+4.6\%}
& \textbf{2.94} \chg{+4.9\%}
& \textbf{1.20} \chg{+2.1\%}
& \textbf{36.16} \chg{+10.3\%}
& \textbf{60.51} \chg{+4.2\%} \\

\bottomrule
\end{tabular}%
}
\end{table*}

Table~\ref{tab:main_results1} summarizes performance across short-term, long-term, and all prediction pairs. DevelopmentODE achieves the best results across all metrics and evaluation settings. On all test pairs, it reduces MSE by $4.89\%$ and improves PCC by $4.24\%$ relative to RNN, with consistent gains in MAE and $R^2$. It also outperforms direct, sequential, and continuous-time baselines. While time-conditioned MLPs and generic Neural ODEs are competitive with or improve upon recurrent models, particularly for long-term transitions, age-conditioned and mixture/stage-based Neural ODEs remain close to the generic Neural ODE and below DevelopmentODE, indicating that continuous-time modeling, age conditioning, or additional dynamical flexibility alone do not explain the gains. The advantage is larger for long-term transitions: DevelopmentODE reduces MSE by $5.12\%$ and improves PCC by $4.61\%$, compared with $4.56\%$ and $3.72\%$ on short-term transitions. Our findings support the benefit of structured developmental dynamics for modeling developmental trajectories.

\subsection{Ablation Study}
\label{sec:ablation}
\begin{wraptable}{r}{0.45\textwidth}
\vspace{-4mm}
\centering
\caption{Ablation study on all test pairs. Relative changes from the full model are shown in parentheses.}
\label{tab:ablation}
\small
\setlength{\tabcolsep}{5pt}
\renewcommand{\arraystretch}{1.06}

\begin{tabular}{lcc}
\toprule
\textbf{Variant}
& \textbf{MSE} $\downarrow$
& \textbf{PCC} $\uparrow$ \\
\midrule

\textbf{Ours}
& \textbf{2.939}
& \textbf{60.51} \\

w/o Canal
& 2.975 {\scriptsize (+1.22\%)}
& 59.96 {\scriptsize (-0.91\%)} \\

w/o Time
& 2.997 {\scriptsize (+1.97\%)}
& 59.58 {\scriptsize (-1.54\%)} \\

w/o Stage
& 2.997 {\scriptsize (+1.97\%)}
& 59.61 {\scriptsize (-1.49\%)} \\

\bottomrule
\end{tabular}
\vspace{-5mm}
\end{wraptable}

We conduct three groups of ablations to examine the key structural choices in
DevelopmentODE. (1) {Developmental geometry}: we remove the population developmental canal
and its associated geometric structure, denoted as \textit{w/o Canal}, to assess
whether population-referenced organization improves dynamical
modeling.
(2) {Temporal information}: we remove temporal conditioning from canal and stage modeling, denoted as
\textit{w/o Time}, to assess whether explicitly modeling when a transition
occurs is necessary for developmental prediction.
(3) {Stage-dependent dynamics}: we remove the ordered stage-specific
modulation, denoted as \textit{w/o Stage}, to evaluate whether developmental
non-stationarity can be adequately captured by a single shared deviation field.
All variants use the same latent representation,
training protocol, and evaluation pairs.
The ablation results consistently support the proposed structural design.
Removing the population canal increases MSE by $1.22\%$ and decreases PCC by
$0.91\%$, indicating that population-referenced developmental geometry provides
a useful inductive structure for individual dynamics. Removing temporal conditioning
produces a larger degradation ($+1.97\%$ MSE and $-1.54\%$ PCC), showing that
developmental dynamics depend on when a transition occurs rather than on the
source state and elapsed interval alone. Removing ordered stage-dependent
modulation yields a comparable degradation ($+1.97\%$ MSE and $-1.49\%$ PCC),
supporting the need to model developmental non-stationarity beyond a single
shared deviation field.

\subsection{Sparsity Analysis}
\begin{wrapfigure}{r}{0.4\textwidth}
    \centering
    \includegraphics[
        width=\linewidth,
        trim=0 0 0 0,
        clip
    ]{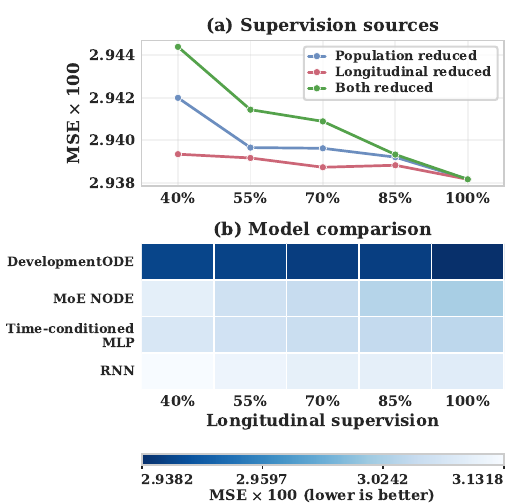}
    \caption{Forecasting error under different amounts of supervision.}
    \label{fig:sparsity}
\end{wrapfigure}
Longitudinal neuroimaging is inherently sparse and unevenly supervised:
population-level observations may span a broad developmental range, while
repeated measurements from the same subject are substantially more limited.
This asymmetry motivates the structured design of DevelopmentODE, which uses
population-level observations to define a shared developmental canal and
reserves longitudinal supervision for learning subject-specific dynamics around
this population structure. We therefore examine the effect of reducing
population and longitudinal supervision separately and jointly across multiple
availability levels. Figure~\ref{fig:sparsity} shows that DevelopmentODE is more
sensitive to reductions in population supervision than in longitudinal
supervision, while remaining stable across the evaluated range and
outperforming representative baselines under reduced longitudinal supervision.
These results support the central design principle of DevelopmentODE: when
subject-specific trajectories are sparsely observed, explicitly separating
population-level developmental structure from individual longitudinal dynamics
provides a useful inductive bias for developmental forecasting.

\subsection{Prediction Horizon Analysis}
\begin{figure*}[!t]
    \centering

    \begin{subfigure}{0.3\textwidth}
        \centering
        \includegraphics[width=\linewidth, trim=0 0 0 0, clip]{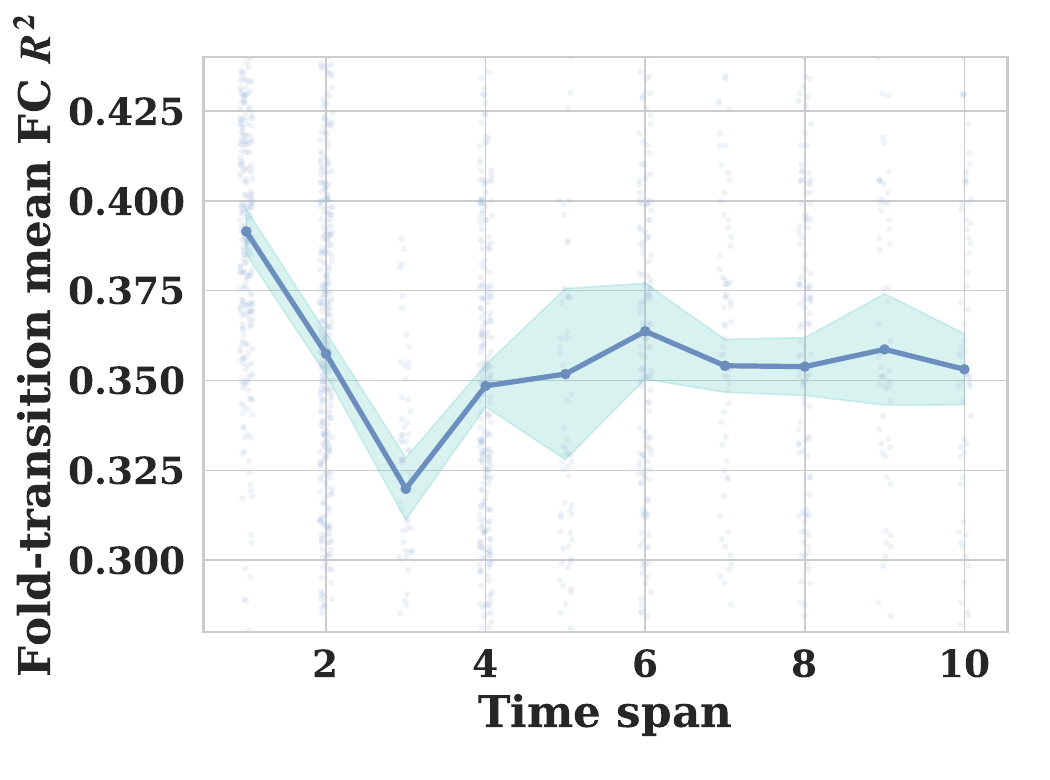}
        \caption{Time-span Effect}
        \label{fig:timespan}
    \end{subfigure}
    \hfill
    \begin{subfigure}{0.3\textwidth}
        \centering
        \includegraphics[width=\linewidth, trim=0 0 0 0, clip]{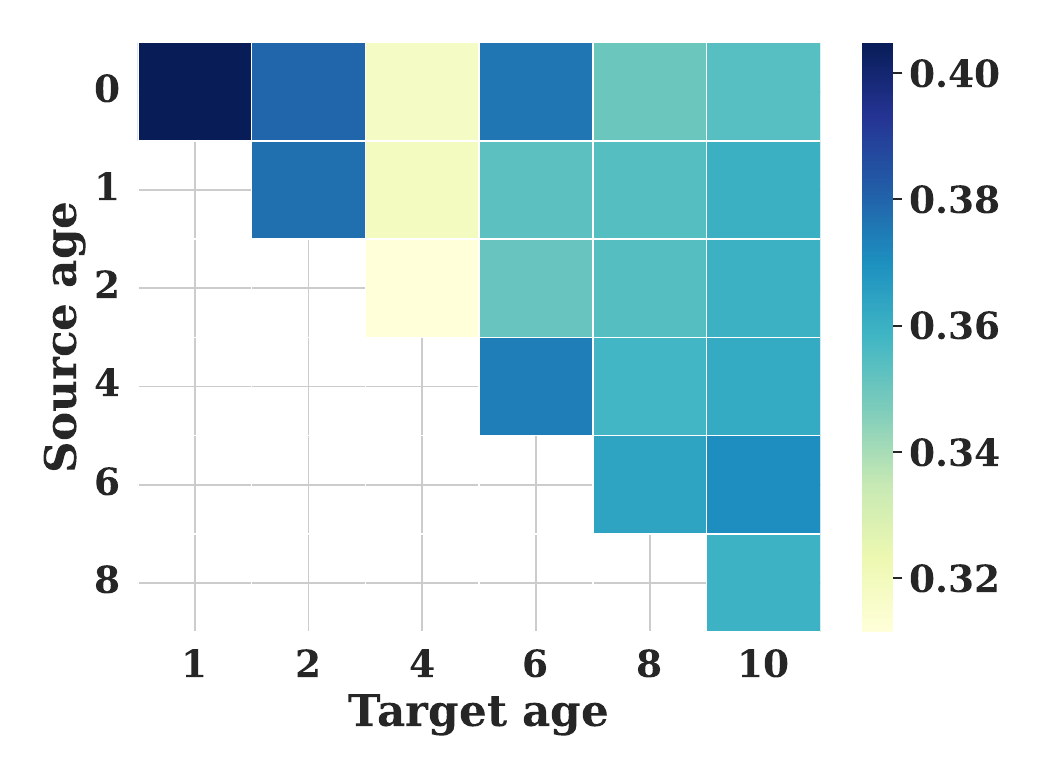}
        \caption{Transition Map}
        \label{fig:st}
    \end{subfigure}
    \hfill
    \begin{subfigure}{0.3\textwidth}
        \centering
        \includegraphics[width=\linewidth, trim=0 0 0 0, clip]{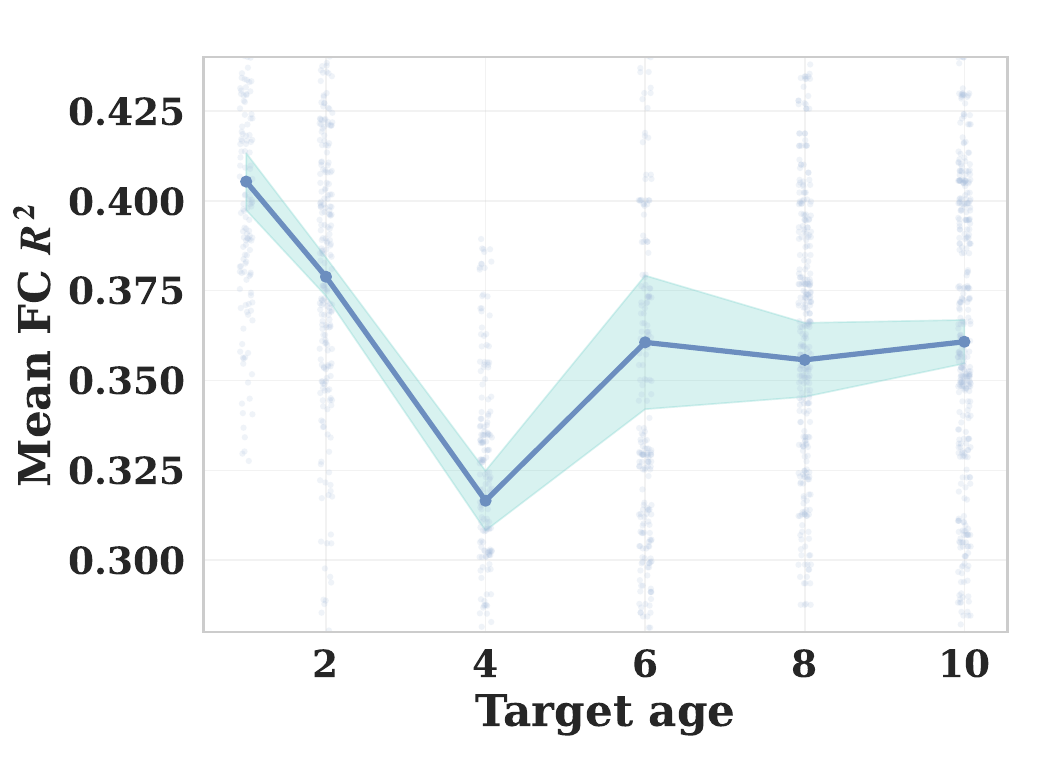}
        \caption{Target-time Effect}
        \label{fig:target}
    \end{subfigure}
\caption{ Temporal analysis of longitudinal FC forecasting. (a) Prediction performance as a function of the source--target temporal interval. (b) Performance across source-time and target-time pairs. (c) Performance across different target time.}
    \label{fig:span}
\end{figure*}

The short-versus-long comparison in Table~\ref{tab:main_results1} summarizes performance using a single temporal threshold, but developmental prediction difficulty may vary continuously with both the elapsed time and the specific source--target transition. We therefore analyze prediction error as a function of the source--target interval $ \Delta t = t_b-t_a $. Figure~\ref{fig:span}(a) shows how forecasting performance varies across prediction horizons. Although longer prediction intervals are generally more difficult, temporal distance alone does not fully characterize prediction difficulty. Figure~\ref{fig:span}(b) reports the source-age--target-age transition map, which preserves the identity of both endpoints rather than collapsing all pairs with the same temporal gap. The heterogeneous pattern across the map shows that forecasting difficulty depends on the developmental transition itself in addition to its duration. 
This provides a more direct diagnostic for the proposed non-stationary developmental formulation: transitions of similar length need not correspond to identical developmental dynamics.

\subsection{Sensitivity analysis}
We conduct additional sensitivity analyses to examine how DevelopmentODE is affected by several implementation choices. Specifically, we vary three factors independently: the numerical ODE solver, the dimensionality of the latent representation, and the placement of developmental stage boundaries, while keeping all other model and optimization settings fixed. The results show that performance is generally stable across different ODE solvers and latent dimensions, whereas the choice of developmental stage boundaries has a more noticeable effect. These analyses suggest that DevelopmentODE is not strongly dependent on several numerical and representation choices, while the stage partition remains an important modeling choice that can influence predictive performance.









\begin{figure*}[!t] \centering \begin{subfigure}{0.3\textwidth} \centering \includegraphics[width=\linewidth]{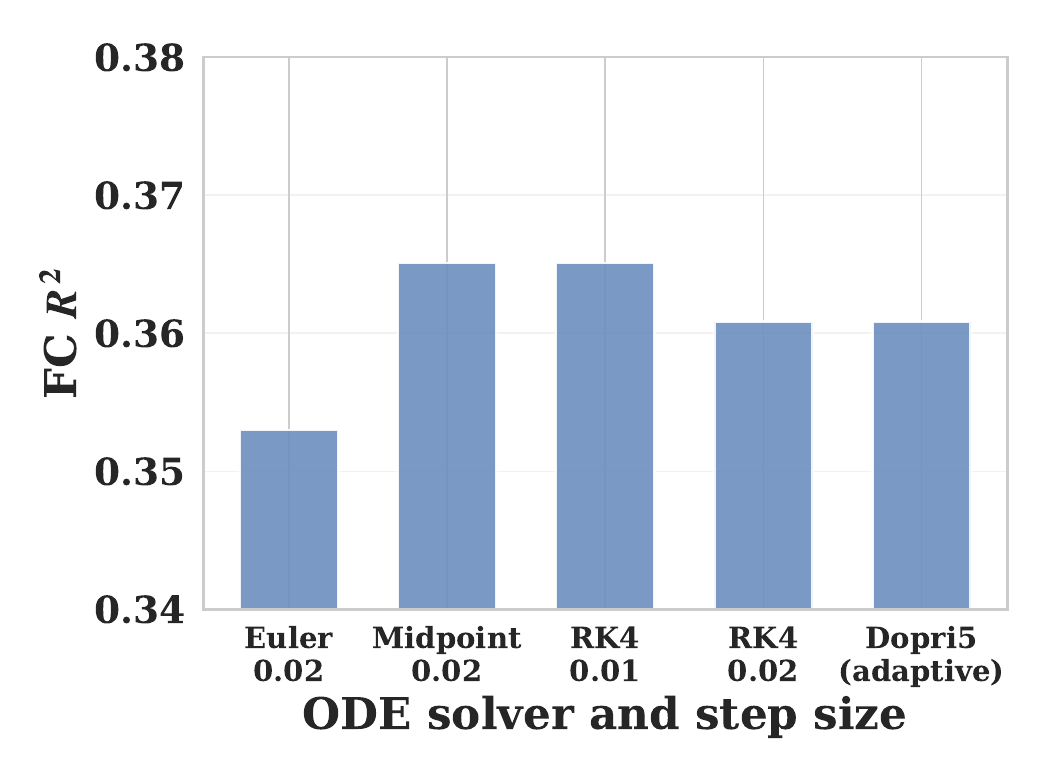} \caption{ODE Solver} \label{fig:solver_robustness} \end{subfigure} \hfill \begin{subfigure}{0.3\textwidth} \centering \includegraphics[width=\linewidth]{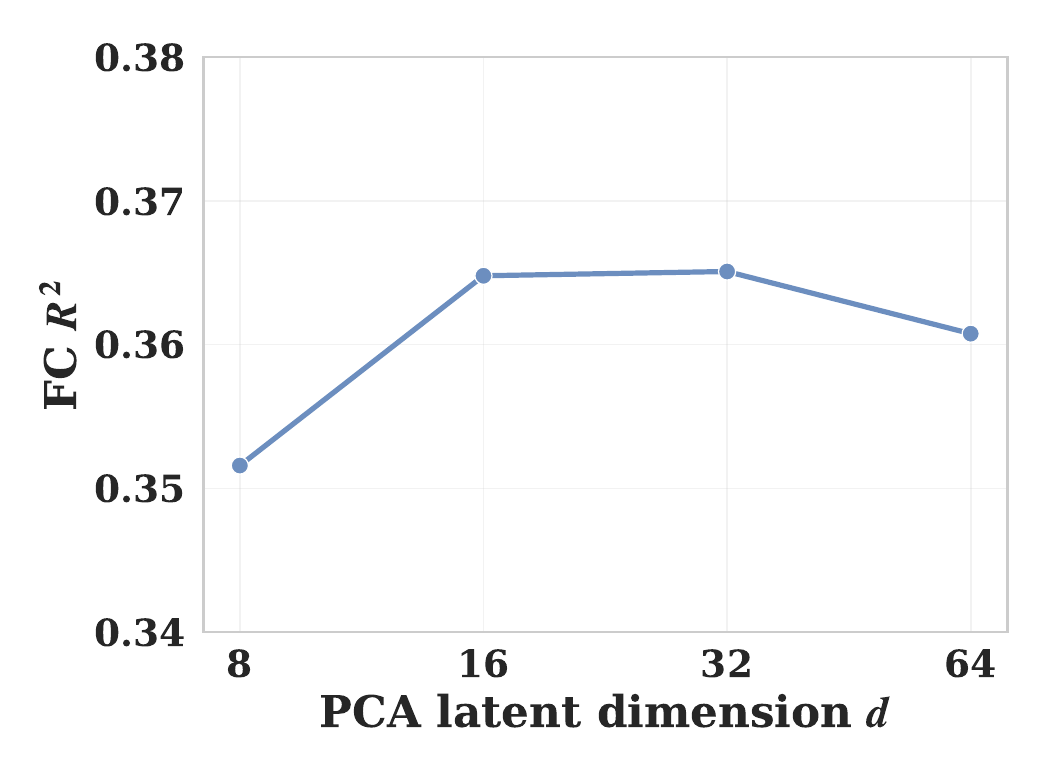} \caption{ Representation Dimension} \label{fig:representation_robustness} \end{subfigure} \hfill \begin{subfigure}{0.3\textwidth} \centering \includegraphics[width=\linewidth]{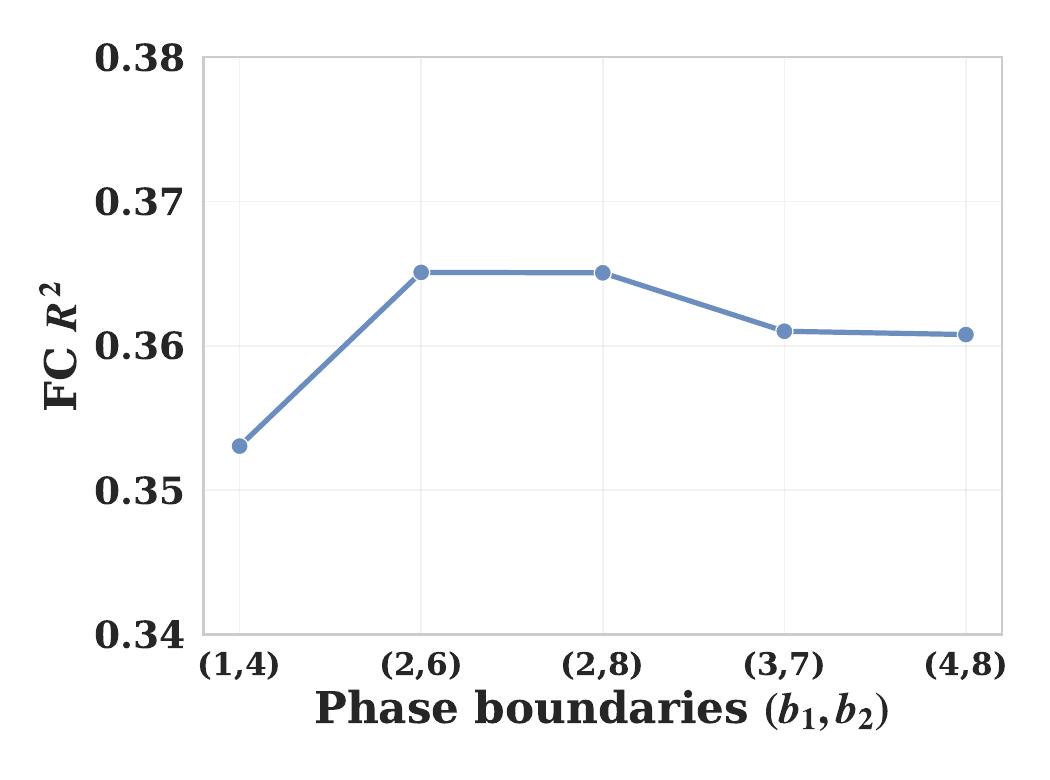} \caption{Stage Boundaries} \label{fig:phase_robustness} \end{subfigure} \caption{ Robustness of DevelopmentODE to numerical solver, latent representation dimension, and developmental stage-boundary choices. } \label{fig:robustness} \end{figure*}

\subsection{Independent Cohort Evaluation}

\begin{wraptable}{r}{0.45\textwidth}
\vspace{-6mm}
\centering
\caption{
Comparison on the external BCP cohort.
MSE and PCC are multiplied by $100$ for readability.
}
\label{tab:external_comparison1}

\small
\setlength{\tabcolsep}{7pt}
\renewcommand{\arraystretch}{1.08}

\begin{tabular}{lcc}
\toprule
\textbf{Model}
& \textbf{MSE} $\downarrow$
& \textbf{PCC} $\uparrow$ \\
\midrule

\textbf{DevelopmentODE}
& \textbf{4.0467}
& \textbf{53.6233} \\

Time-MLP
& 4.6256
& 48.6872 \\

Direct MLP
& 4.6728
& 48.3735 \\

Age-NODE
& 5.0071
& 44.4017 \\

Generic NODE
& 5.0102
& 44.3411 \\

\bottomrule
\end{tabular}

\vspace{-4mm}
\end{wraptable}

To further evaluate the general applicability of the proposed framework,
we independently train and evaluate all methods on the BCP cohort using the
same experimental protocol. Table~\ref{tab:external_comparison1} reports performance on the external cohort using the same evaluation metrics.
DevelopmentODE consistently achieves lower prediction error and higher
structural agreement than the competing methods. The results indicate that
the proposed developmental modeling strategy remains effective on an
independently collected cohort. Please refer to Appendix~\ref{Indepent} for the full results on BCP.
\section{Conclusion}

We introduced DevelopmentODE, a structured continuous-time framework for modeling sparse and irregular longitudinal data. Rather than learning individual dynamics in an unconstrained latent space, DevelopmentODE organizes them on a developmental geometry that integrates shared population-level structure with subject-specific variation. Across internal and external evaluations, the proposed formulation consistently improves over discrete-time, generic Neural ODE, and age-conditioned alternatives. These results highlight the value of incorporating structured population information into continuous-time models and suggest a general direction for learning reliable dynamics from sparse longitudinal observations.

\section*{AI Use Statement}
Generative AI tools were used to assist with language polishing, improving clarity and readability, refining the presentation and organization of parts of the manuscript, and assisting with minor formatting tasks such as LaTeX and table formatting. The authors reviewed and revised all AI-assisted text and retained full control over the scientific content, methodology, experiments, analyses, and conclusions. Generative AI tools were not used to generate experimental data or results. The authors take full responsibility for the final content of the paper, including any text or artifacts produced with the assistance of generative AI.

\section*{Ethics Statement}
This study uses previously collected longitudinal neuroimaging data from the
Early Brain Development Study (EBDS) and the Baby Connectome Project (BCP).
Our work involves secondary analysis of these existing datasets and does not
include new participant recruitment, data collection, or intervention.
All analyses were conducted in accordance with the data-use requirements of
the respective datasets. The present study focuses on methodological
development for longitudinal prediction and does not make clinical decisions
or provide subject-level diagnostic conclusions.

\bibliography{iclr2026_conference}
\bibliographystyle{iclr2026_conference}

\appendix

\section{Dataset Details}
\label{dataset}

\subsection{Longitudinal Prediction Cohort}

Following the previous preprocessing pipeline, we use data from the University of North Carolina Early Brain Development
Study (EBDS), a longitudinal neuroimaging cohort spanning the first decade of
life~\citep{chen2026functional}. After preprocessing and quality control, the dataset contains
1,476 resting-state functional MRI (rsfMRI) scans from 633 children.
Measurements were collected at approximately seven developmental ages:
3 weeks, 1, 2, 4, 6, 8, and 10 years.
The numbers of usable scans at these ages are 372, 273, 220, 125, 175, 155,
and 156, respectively. The resulting dataset is longitudinal but
substantially irregular: subjects differ in both the number of available
visits and the intervals between consecutive observations. This setting
directly motivates our continuous-time formulation.
For twin pairs, we retain only one child from each pair to reduce dependence
between subjects. The retained child is selected as the one with the largest
number of high-quality rsfMRI scans across developmental ages. Controls satisfy the
following criteria: gestational age of at least 32 weeks, no significant
medical or neurological conditions, no maternal psychiatric diagnosis, and
no neonatal illness requiring more than 24 hours of neonatal intensive care.

Because EBDS spans multiple years of data collection, scans were acquired
using several Siemens 3T MRI systems. Data from ages 0--6 years and a subset
of the 8--10 year measurements were acquired on Siemens Allegra or Tim Trio
scanners, whereas most scans at ages 8--10 years were acquired on a Siemens
Magnetom Prisma scanner.
Children between 0 and 2 years of age were scanned during natural sleep.
Older children were scanned awake under age-appropriate viewing conditions,
including cartoons, movies, or fixation. Scanner- and acquisition-related
effects are accounted for during preprocessing and feature harmonization, as
described below.
All rsfMRI scans are processed using a standardized pipeline based on FSL,
ANTs, and AFNI. Processing includes rigid-body motion correction,
registration, temporal band-pass filtering, motion scrubbing, nuisance
regression, and global signal regression. Scans with fewer than three
minutes of usable data after motion scrubbing are excluded. Functional images are first aligned to an age-specific template and are
subsequently transformed into a common 2-year template space.

\subsection{Longitudinal Prediction Protocol}
\label{app:data_split}

Our main experiments focus on within-subject longitudinal prediction. For a
subject with observations at developmental ages
$t_i^1 < \cdots < t_i^{n_i}$, we construct forward prediction pairs
$(t_a,t_b)$ with $t_b>t_a$. The observation at $t_a$ is used as the source
state and the same subject's observation at $t_b$ serves as the prediction
target. We do not construct longitudinal targets by pairing observations from
different subjects.
Data partitioning is performed at the \emph{subject level}, such that all
observations from the same child belong to a single split. This prevents
information leakage between source and target observations of the same
subject. Because visit intervals vary substantially across subjects, the
resulting evaluation includes both short- and long-horizon developmental
transitions. This protocol evaluates whether a model can infer an
individual's future developmental state from a sparse observation while
remaining defined for irregular and arbitrary prediction intervals.

\subsection{BCP auxiliary cohort}
We additionally use an independent BCP cohort for auxiliary validation. The processed dataset contains 121 resting-state fMRI observations from children in the 0--5 year age range, with 25, 35, 34, 13, 8, and 6 observations in the six age groups, respectively. Each valid rsfMRI run is processed independently by discarding the first five volumes, extracting regional time series with 100 cortical parcels and seven functional networks, applying 6-mm FWHM spatial smoothing, linear detrending, 0.01--0.10 Hz band-pass filtering, and sample-wise $z$-score normalization. When multiple valid runs are available for the same subject and age, their independently standardized ROI time series are concatenated along the temporal dimension, after which a $100\times100$ functional connectivity matrix is constructed using pairwise Pearson correlations. One time point with severe atlas coverage failure is excluded. All retained matrices are checked for dimensional consistency, finite values, symmetry, valid correlation range, and consistent ROI ordering.

\section{Implementation Details}
\label{sec:implementation}

DevelopmentODE is implemented under a strictly fold-aware longitudinal
prediction protocol. Subjects are split into training, validation, and test
sets before any source--target pairs are constructed, ensuring that all scans
from the same subject remain within a single split. For each fold, we
independently fit the latent representation, age normalization, population
trajectory, model parameters, and validation checkpoint using only the
corresponding training subjects.
For each scan, the functional connectivity (FC) matrix is thresholded at
$0.1$, vectorized, and projected into a $32$-dimensional PCA space~\citep{article12,Rubinov2010ComplexNM}. PCA is
used only to define the state space in which the developmental dynamics are
modeled, while prediction performance is optimized and evaluated after
mapping the predicted state back to the original FC space. The PCA
transformation is fitted exclusively on training scans within each fold.
Developmental ages are normalized independently within each fold using the
minimum and maximum ages observed among its training subjects.
Visits from each subject are sorted chronologically before longitudinal
transitions are constructed. In the \texttt{all} setting, we use every
strictly forward source--target pair available for each subject, whereas the
\texttt{adjacent} setting uses only consecutive visits.

We represent population-level development using a differentiable
developmental canal and model subject-specific dynamics relative to this
shared trajectory. The canal is implemented as a small MLP.
Before longitudinal training, the canal is initialized from population
statistics computed only from the fold-training set. Training scans are
grouped by their rounded acquisition ages, and we compute the mean PCA
representation within each age group. We first fit the output layer of the
canal to these age-specific population means using least squares over the
hidden representation. We then refine the complete canal for $500$ AdamW
steps using the discrepancy between the predicted canal states and the
corresponding age-specific population means. This stage provides a smooth
population developmental trajectory before subject-level longitudinal
transitions are introduced.

Longitudinal prediction starts from the observed state at the source age and
integrates the learned developmental dynamics until the target age. The full
DevelopmentODE combines three components within this evolution: the change of
the population canal over age, a shared nonlinear deviation field that models
subject-specific motion around the canal, and an age-dependent
component.
The change of the population canal is obtained directly through automatic
differentiation of the canal network with respect to age. 
The full model uses three smoothly ordered
developmental stages. The two stage transitions are centered at normalized
ages $0.2$ and $0.6$ and are connected using sigmoid gates with temperature
$0.10$.

All DevelopmentODE variants and learning-based baselines follow the same
optimization protocol. To avoid over-representing subjects with more
longitudinal observations, mini-batches are sampled in a subject-balanced
manner: each update first samples eligible subjects and then selects one valid
forward transition from each sampled subject.

Unless otherwise stated, models are optimized using AdamW with learning rate
$5\times10^{-4}$, weight decay $10^{-4}$, batch size $256$, and gradient
clipping at $1.0$. Training is stopped after $20$ epochs without improvement
in validation FC-space MSE. The primary training objective is the
reconstruction error between the predicted and observed target FC matrices
after mapping predictions back from the PCA space. We additionally use the
centering objective described in the main method, weighted by $0.05$, to
preserve the interpretation of the learned population canal during
longitudinal optimization. Except for sparsity experiments that
explicitly freeze the pretrained canal, the population canal and
subject-specific developmental dynamics are jointly refined during
longitudinal training.

Numerical integration is performed using \texttt{torchdiffeq.odeint}. We use
fixed-step fourth-order Runge--Kutta (RK4) as the default solver with an
age-normalized step size of $0.02$. For efficient batching, every
source--target transition is internally mapped to the same unit integration
interval. During integration, the vector field is scaled by the actual
normalized source--target age difference, so transitions of different lengths
share the same numerical interface while preserving their true developmental
duration. We additionally evaluate the adaptive \texttt{dopri5} solver in the
solver-robustness experiments.

Checkpoints are selected exclusively according to validation FC-space MSE.
After restoring the best validation checkpoint, the corresponding held-out
test subjects are evaluated without further model selection.
Short-term transitions are defined as those spanning at most $0.2$ in
normalized age units, and all remaining transitions are assigned to the
long-term subset.

All baselines use the same fold-specific PCA representation, subject splits,
training pairs, FC-space objective, optimizer, and checkpoint-selection rule.
The default hidden width is 64. Direct MLP uses the source latent state and
elapsed interval, whereas Time-conditioned MLP additionally receives source
and target ages. RNN, GRU, and LSTM are single-layer models operating on a
two-token source--target representation. The Transformer contains one encoder
layer, four attention heads, and a feed-forward width of 256. Generic and
age-conditioned Neural ODEs use a two-layer $\tanh$ vector field; the latter
additionally receive absolute developmental age. MoE Neural ODE uses three
vector fields with a learned soft age-dependent router, while Stage Neural ODE
uses three hard-routed fields with fixed boundaries $(0.2,0.6)$. Output layers
are initialized to zero where applicable, giving an initial identity
prediction and providing a common stable starting point.

For every source--target pair, MSE, MAE, $R^2$, and PCC were first computed
over all FC entries and then averaged over pairs. Short-term predictions were
defined by $\Delta t\leq0.2$ in normalized age (equivalent to at most two years
on the ten-year scale), and long-term predictions by $\Delta t>0.2$. In the tables, MSE, $R^2$, and
PCC are multiplied by 100; MAE is multiplied by 10. Test data were not used for
optimization or checkpoint selection.

The models were implemented in PyTorch, and differentiable ODE integration was
performed with \texttt{torchdiffeq}. Random seeds were applied to Python,
NumPy, and PyTorch, and deterministic cuDNN execution was enabled. Each run
stored its subject IDs, hyperparameters, PCA state, validation history, selected
checkpoint, and pair- and subject-averaged test metrics.


\section{Additional Experiment And Analysis}

\subsection{Developmental Geometry Across Age}
\label{app:developmental_geometry}

DevelopmentODE operates in a 32-dimensional latent space and organizes
developmental dynamics around a smooth population canal $\mu_\eta(t)$.
For each subject, individual variation is represented relative to this shared
trajectory as $r_i(t)=z_i(t)-\mu_\eta(t)$, with $\tilde r_i(t)$ denoting its
component in the complementary developmental geometry. The deviation dynamics
combine three components: a shared nonlinear field $\tilde f_\theta$ that
captures structure pooled across subjects and developmental ages, an
age-dependent contraction component $\lambda(t)$ that modulates the evolution
of subject-specific deviations, and low-rank stage-specific deformation fields
$A_k$ that provide additional age-dependent directional deformations of the
shared dynamics. In particular, the ordered stage weights $\pi_k(t)$ define
$
\lambda(t)=\sum_{k=1}^{M}\pi_k(t)\lambda_k,
$
where the positive coefficients $\lambda_k$ are learned during longitudinal
training. The corresponding contraction term is
$-\lambda(t)\tilde r_i(t)$. The low-rank corrections share a common
low-dimensional projection and use stage-specific modulation and bounded gates,
so that their age-dependent contribution can be written at a high level as
$\sum_k \pi_k(t)A_k(\tilde r_i(t))$. Thus, these implementation-level
components instantiate the ordered age-dependent deformations
$\Delta\tilde f_k$ used in the main formulation, with each deformation
combining a linear component and a capacity-controlled directional
correction. The canal is
initialized from age-specific population statistics and subsequently optimized
jointly with the deviation dynamics. Together, these components implement the
central design of DevelopmentODE by assigning population progression,
subject-specific variation, and age-dependent changes in the dynamics to
distinct but coupled components of the continuous-time model.

We next examine the resulting quantities across developmental age.
Population-level progression is characterized by the magnitude of the canal
velocity, $\|\dot{\mu}_\eta(t)\|_2$, while subject-specific variation is
summarized by the magnitude of the projected deviation,
$\|\tilde r_i(t)\|_2$. We additionally report $\lambda(t)$, the
age-dependent contraction coefficient applied directly to the restoring
component $-\lambda(t)\tilde r_i(t)$ of the deviation dynamics. Because
$\lambda(t)$ is obtained by smoothly combining the learned stage-specific
coefficients $\lambda_k$ using the ordered age weights $\pi_k(t)$, its profile
reflects both these learned coefficients and the predefined smooth stage
organization. The stage-specific fields $A_k$ provide complementary
directional corrections through which the shared dynamics can vary across
developmental stages. These quantities characterize different components of
the fitted dynamical model and are not assumed to correspond directly to any
single empirical measure or biological mechanism of longitudinal change. All
reported statistics are computed on held-out subjects and averaged over the
five subject-disjoint folds. Table~\ref{tab:developmental_geometry} reports
the resulting patterns across age, allowing us to examine how population
progression, individual variation, and age-dependent contraction are organized
within the fitted developmental geometry.

As shown in Table~\ref{tab:developmental_geometry}, the fitted population
trajectory has its largest canal velocity at 3 weeks, with
$\|\dot{\mu}_{\eta}(t)\|_2=284.007$, and the magnitude decreases across the
evaluated ages to $33.623$ at 10 years. The projected individual deviation
shows a different age-dependent profile: its average magnitude increases from
$11.361$ at 3 weeks to $13.882$ at 2 years, decreases at the intermediate
ages, and reaches $14.072$ at 10 years. These different profiles are
consistent with the intended separation between shared population progression
and variation around the developmental canal. They should be interpreted as
properties of the fitted representation rather than as direct estimates of
underlying neurodevelopmental mechanisms.

\begin{table}[t]
\centering
\caption{
Developmental geometry across age on held-out subjects.
Values are averaged over five subject-disjoint folds.
}
\label{tab:developmental_geometry}
\small
\setlength{\tabcolsep}{5pt}
\renewcommand{\arraystretch}{1.05}
\begin{tabular}{lrrrr}
\toprule
\textbf{Age}
& \textbf{N}
& $\|\dot{\mu}\|_2$
& $\lambda(t)$
& $\mathbb{E}\|\tilde r\|_2$ \\
\midrule
3 weeks  & 228 & 284.007 & 27.782 & 11.361 \\
1 year   & 233 & 109.156 & 25.219 & 13.384 \\
2 years  & 199 & 101.897 & 21.251 & 13.882 \\
4 years  & 116 & 73.443  & 14.464 & 12.672 \\
6 years  & 156 & 52.219  & 11.598 & 12.197 \\
8 years  & 152 & 37.022  & 10.192 & 13.739 \\
10 years & 143 & 33.623  & 9.851  & 14.072 \\
\bottomrule
\end{tabular}
\end{table}

\subsection{Population and Individual Components of Longitudinal Development}
\label{app:population_individual_composition}

We next examine how observed within-subject developmental changes are
distributed relative to the learned population geometry. For each held-out
longitudinal pair $(t_a,t_b)$, we compute the observed latent displacement
$\Delta z_i=z_i(t_b)-z_i(t_a)$ and decompose it with respect to the population
developmental tangent evaluated at the midpoint of the interval. We report the
fraction of displacement energy aligned with the population direction and the
fraction contained in its complementary space; the two fractions therefore sum
to one.
Table~\ref{tab:transition_composition} shows substantial heterogeneity across
developmental transitions. Early transitions contain relatively large
population-aligned components, including $0.530$ for
3 weeks$\rightarrow$1 year and $0.489$ for
3 weeks$\rightarrow$2 years. For many other transitions, however, most of the
observed displacement lies in the complementary space. This indicates that
longitudinal development contains both shared population progression and
substantial variation outside the instantaneous population direction, with
their relative contributions changing across developmental intervals.

\begin{table}[t]
\centering
\caption{
Population-aligned and complementary components of observed longitudinal
changes on held-out subjects.
}
\label{tab:transition_composition}
\scriptsize
\setlength{\tabcolsep}{5pt}
\renewcommand{\arraystretch}{1.03}
\begin{tabular}{lrrr}
\toprule
\textbf{Transition}
& \textbf{N}
& \textbf{Population}
& \textbf{Deviation} \\
\midrule
3 weeks $\rightarrow$ 1 year   & 159 & 0.530 & 0.470 \\
3 weeks $\rightarrow$ 2 years  & 109 & 0.489 & 0.511 \\
3 weeks $\rightarrow$ 4 years  & 71  & 0.022 & 0.978 \\
3 weeks $\rightarrow$ 6 years  & 65  & 0.033 & 0.967 \\
3 weeks $\rightarrow$ 8 years  & 90  & 0.040 & 0.960 \\
3 weeks $\rightarrow$ 10 years & 76  & 0.106 & 0.894 \\
1 year $\rightarrow$ 2 years   & 130 & 0.176 & 0.824 \\
1 year $\rightarrow$ 4 years   & 66  & 0.438 & 0.562 \\
1 year $\rightarrow$ 6 years   & 78  & 0.159 & 0.841 \\
1 year $\rightarrow$ 8 years   & 81  & 0.043 & 0.957 \\
1 year $\rightarrow$ 10 years  & 72  & 0.037 & 0.963 \\
2 years $\rightarrow$ 4 years  & 54  & 0.372 & 0.628 \\
2 years $\rightarrow$ 6 years  & 82  & 0.259 & 0.741 \\
2 years $\rightarrow$ 8 years  & 71  & 0.031 & 0.969 \\
2 years $\rightarrow$ 10 years & 61  & 0.023 & 0.977 \\
4 years $\rightarrow$ 6 years  & 53  & 0.076 & 0.924 \\
4 years $\rightarrow$ 8 years  & 60  & 0.101 & 0.899 \\
4 years $\rightarrow$ 10 years & 57  & 0.151 & 0.849 \\
6 years $\rightarrow$ 8 years  & 82  & 0.143 & 0.857 \\
6 years $\rightarrow$ 10 years & 82  & 0.138 & 0.862 \\
8 years $\rightarrow$ 10 years & 87  & 0.092 & 0.908 \\
\bottomrule
\end{tabular}
\end{table}

\paragraph{Age-dependent modulation and deviation persistence.}
We further compare the learned age-dependent modulation with the empirical
persistence of individual deviations. For each held-out adjacent transition,
deviation persistence is measured by
$\log(\|\tilde r_i(t_b)\|_2/\|\tilde r_i(t_a)\|_2)$, where negative values
indicate decreasing deviation magnitude and positive values indicate increasing
deviation magnitude. Table~\ref{tab:lambda_persistence} shows that the learned
modulation decreases smoothly across development, whereas the observed
deviation magnitude changes non-monotonically. Deviations decrease most clearly
over the 2$\rightarrow$4 and 4$\rightarrow$6 year transitions, while they
increase over 3 weeks$\rightarrow$1 year and 6$\rightarrow$8 years. We
therefore interpret $\lambda(t)$ as an age-dependent dynamical modulation of
individual variation rather than a direct proxy for the observed change in
deviation magnitude.

\begin{table}[t]
\centering
\caption{
Learned age-dependent modulation and observed deviation persistence for
adjacent developmental transitions.
}
\label{tab:lambda_persistence}
\small
\setlength{\tabcolsep}{5pt}
\renewcommand{\arraystretch}{1.05}
\begin{tabular}{lrrr}
\toprule
\textbf{Transition}
& \textbf{N}
& $\bar{\lambda}$
& $\log \frac{\|\tilde r_b\|_2}{\|\tilde r_a\|_2}$ \\
\midrule
3 weeks $\rightarrow$ 1 year   & 159 & 26.693 &  0.181 \\
1 year $\rightarrow$ 2 years   & 130 & 23.344 &  0.038 \\
2 years $\rightarrow$ 4 years  & 54  & 17.215 & -0.073 \\
4 years $\rightarrow$ 6 years  & 53  & 12.813 & -0.056 \\
6 years $\rightarrow$ 8 years  & 82  & 10.715 &  0.106 \\
8 years $\rightarrow$ 10 years & 87  & 9.942  &  0.004 \\
\bottomrule
\end{tabular}
\end{table}

\subsection{Training on Adjacent Longitudinal Transitions}
\label{sec:adjacent_training}

We further consider a more restrictive training setting in which models are
trained only on adjacent within-subject transitions, while evaluation is
performed on all forward test pairs. This setting examines whether modeling
developmental dynamics from consecutive observations remains effective when
predictions are evaluated over both short- and longer-range transitions.
As shown in Table~\ref{taba}, DevelopmentODE achieves the
best performance across all metrics and evaluation horizons. On all test
pairs, it reduces MSE by $6.09\%$ relative
to the RNN baseline. The improvement is larger for long-term transitions,
where DevelopmentODE reduces MSE by $7.36\%$, compared with $4.13\%$ on short-term transitions,
respectively. DevelopmentODE also consistently outperforms the generic,
age-conditioned, mixture, and stage-based Neural ODE alternatives under this
training protocol. These results show that the advantage of the proposed
structured developmental dynamics remains evident when training is restricted
to consecutive longitudinal observations.
\begin{table*}[t!]
\caption{
Evaluation results when models are trained only on adjacent within-subject
transitions and evaluated on all forward test pairs.
Metrics include MSE ($\downarrow$), MAE ($\downarrow$), $R^2$ ($\uparrow$),
and PCC ($\uparrow$).
The change rate (\%) is computed relative to the RNN baseline.
The best results are in \textbf{bold}, and the second-best are
\underline{underlined}.
}
\label{taba}
\centering
\scriptsize
\setlength{\tabcolsep}{2.0pt}
\renewcommand{\arraystretch}{1.05}
\resizebox{\textwidth}{!}{%
\begin{tabular}{@{}l*{24}{c}@{}}
\toprule
\textbf{Model}
& \multicolumn{8}{c}{\textbf{Short-term}}
& \multicolumn{8}{c}{\textbf{Long-term}}
& \multicolumn{8}{c}{\textbf{All}} \\
\cmidrule(lr){2-9}
\cmidrule(lr){10-17}
\cmidrule(lr){18-25}
& MSE & $\downarrow$ & MAE & $\downarrow$ & $R^2$ & $\uparrow$ & PCC & $\uparrow$
& MSE & $\downarrow$ & MAE & $\downarrow$ & $R^2$ & $\uparrow$ & PCC & $\uparrow$
& MSE & $\downarrow$ & MAE & $\downarrow$ & $R^2$ & $\uparrow$ & PCC & $\uparrow$ \\
\midrule
RNN & 3.06 & -- & 1.22 & -- & 34.78 & -- & 59.61 & -- & 3.17 & -- & 1.24 & -- & 29.96 & -- & 56.00 & -- & 3.13 & -- & 1.23 & -- & 31.88 & -- & 57.44 & -- \\
GRU & 3.06 & -0.09 & 1.22 & 0.13 & 34.75 & -0.09 & 59.59 & -0.02 & 3.20 & -0.64 & 1.24 & -0.21 & 29.51 & -1.48 & 55.71 & -0.53 & 3.14 & -0.42 & 1.23 & -0.07 & 31.60 & -0.87 & 57.26 & -0.32 \\
LSTM & 3.06 & 0.01 & 1.22 & 0.24 & 34.81 & 0.10 & 59.63 & 0.03 & 3.18 & -0.21 & 1.24 & 0.09 & 29.81 & -0.49 & 55.93 & -0.13 & 3.13 & -0.12 & 1.23 & 0.15 & 31.81 & -0.22 & 57.41 & -0.06 \\
Transformer & 3.04 & 0.79 & 1.22 & 0.24 & 35.28 & 1.46 & 60.01 & 0.68 & 3.06 & 3.69 & 1.21 & 2.02 & 32.63 & 8.94 & 57.86 & 3.31 & 3.05 & 2.56 & 1.22 & 1.32 & 33.69 & 5.68 & 58.72 & 2.23 \\
Direct MLP & 3.01 & 1.66 & 1.22 & 0.35 & 35.92 & 3.28 & 60.36 & 1.26 & 3.13 & 1.53 & 1.24 & 0.22 & 31.07 & 3.73 & 56.70 & 1.24 & 3.08 & 1.58 & 1.23 & 0.27 & 33.01 & 3.54 & 58.16 & 1.25 \\
Time-conditioned MLP & \underline{3.01} & 1.75 & \underline{1.22} & 0.37 & \underline{35.97} & 3.44 & \underline{60.39} & 1.31 & 3.11 & 1.98 & 1.23 & 0.52 & 31.40 & 4.81 & 56.95 & 1.68 & 3.07 & 1.90 & 1.23 & 0.46 & 33.22 & 4.22 & 58.32 & 1.54 \\
Generic Neural ODE & 3.12 & -1.91 & 1.23 & -0.56 & 33.65 & -3.24 & 58.69 & -1.54 & 3.01 & 5.11 & 1.21 & 2.26 & 33.62 & 12.23 & 58.60 & 4.64 & 3.06 & 2.36 & 1.22 & 1.14 & 33.63 & 5.48 & 58.63 & 2.07 \\
Age-conditioned Neural ODE & 3.12 & -1.89 & 1.23 & -0.55 & 33.66 & -3.21 & 58.69 & -1.53 & 3.01 & 5.16 & 1.21 & 2.35 & 33.64 & 12.31 & 58.64 & 4.70 & 3.06 & 2.40 & 1.22 & 1.20 & 33.64 & 5.53 & 58.65 & 2.11 \\
MoE Neural ODE & 3.11 & -1.42 & 1.23 & -0.46 & 33.96 & -2.35 & 58.92 & -1.15 & \underline{3.00} & 5.44 & 1.21 & 2.38 & \underline{33.85} & 12.99 & \underline{58.78} & 4.96 & \underline{3.04} & 2.75 & 1.22 & 1.25 & \underline{33.89} & 6.30 & \underline{58.83} & 2.42 \\
Stage Neural ODE & 3.10 & -1.30 & 1.22 & -0.31 & 33.99 & -2.25 & 59.06 & -0.93 & 3.04 & 4.26 & \underline{1.21} & 2.42 & 32.99 & 10.12 & 58.22 & 3.96 & 3.07 & 2.08 & \underline{1.22} & 1.34 & 33.39 & 4.73 & 58.55 & 1.93 \\
\midrule
DevelopmentODE (ours) & \textbf{2.94} & 4.13 & \textbf{1.21} & 1.10 & \textbf{37.49} & 7.80 & \textbf{61.55} & 3.25 & \textbf{2.94} & 7.36 & \textbf{1.20} & 3.45 & \textbf{35.25} & 17.69 & \textbf{59.82} & 6.81 & \textbf{2.94} & 6.09 & \textbf{1.20} & 2.51 & \textbf{36.14} & 13.37 & \textbf{60.50} & 5.33 \\
\bottomrule
\end{tabular}%
}
\end{table*}
\label{sec:richer_stage}
\subsection{Main Table with Standard Deviation}
To provide additional information about variation, we report the standard deviation for each metric in Table~\ref{tab:main_results}. The reported SD values are computed for the main evaluation.
\begin{table*}[t!]
\caption{Evaluation results on short-term, long-term, and all prediction pairs. Metrics include MSE ($\downarrow$), MAE ($\downarrow$), $R^2$ ($\uparrow$), and PCC ($\uparrow$). The SD columns report the standard deviation. The best results are in \textbf{bold}, and the second-best are \underline{underlined}.}
\label{tab:main_results}
\centering
\scriptsize
\setlength{\tabcolsep}{2.0pt}
\renewcommand{\arraystretch}{1.05}
\resizebox{\textwidth}{!}{%
\begin{tabular}{@{}l*{24}{c}@{}}
\toprule
\textbf{Model}
& \multicolumn{8}{c}{\textbf{Short-term}}
& \multicolumn{8}{c}{\textbf{Long-term}}
& \multicolumn{8}{c}{\textbf{All}} \\
\cmidrule(lr){2-9}
\cmidrule(lr){10-17}
\cmidrule(lr){18-25}
& MSE & SD
& MAE & SD
& $R^2$ & SD
& PCC & SD
& MSE & SD
& MAE & SD
& $R^2$ & SD
& PCC & SD
& MSE & SD
& MAE & SD
& $R^2$ & SD
& PCC & SD \\
\midrule
RNN
& 3.07 & 0.064 & 1.23 & 0.011 & 34.51 & 0.488 & 59.36 & 0.354%
& 3.10 & 0.040 & 1.23 & 0.008 & 31.65 & 0.804 & 57.19 & 0.585%
& 3.09 & 0.039 & 1.23 & 0.007 & 32.80 & 0.562 & 58.05 & 0.395 \\
GRU
& 3.09 & 0.064 & 1.23 & 0.010 & 34.29 & 0.510 & 59.18 & 0.379%
& 3.10 & 0.037 & 1.23 & 0.004 & 31.64 & 0.723 & 57.18 & 0.531%
& 3.10 & 0.036 & 1.23 & 0.004 & 32.69 & 0.483 & 57.97 & 0.342 \\
LSTM
& 3.07 & 0.067 & 1.22 & 0.010 & 34.59 & 0.450 & 59.40 & 0.346%
& 3.09 & 0.035 & 1.22 & 0.005 & 31.96 & 0.881 & 57.43 & 0.628%
& 3.08 & 0.037 & 1.22 & 0.005 & 33.01 & 0.547 & 58.22 & 0.381 \\
Transformer
& 3.04 & 0.070 & \underline{1.22} & 0.011 & 35.32 & 0.530 & 60.02 & 0.387%
& 3.02 & 0.036 & 1.21 & 0.007 & 33.42 & 0.854 & 58.47 & 0.601%
& 3.03 & 0.040 & 1.21 & 0.007 & 34.17 & 0.607 & 59.09 & 0.408 \\
Direct MLP
& 3.03 & 0.062 & 1.22 & 0.008 & 35.58 & 0.421 & 60.09 & 0.317%
& 3.07 & 0.035 & 1.23 & 0.005 & 32.41 & 0.806 & 57.66 & 0.583%
& 3.05 & 0.038 & 1.23 & 0.004 & 33.68 & 0.545 & 58.63 & 0.401 \\
Time-conditioned MLP
& \underline{3.02} & 0.062 & 1.22 & 0.011 & \underline{35.67} & 0.518 & \underline{60.16} & 0.388%
& 3.05 & 0.034 & 1.22 & 0.006 & 32.80 & 0.887 & 57.98 & 0.638%
& 3.04 & 0.039 & 1.22 & 0.007 & 33.95 & 0.632 & 58.85 & 0.452 \\
Generic Neural ODE
& 3.08 & 0.067 & 1.22 & 0.011 & 34.45 & 0.499 & 59.26 & 0.417%
& 2.98 & 0.033 & 1.20 & 0.007 & 34.31 & 0.817 & 59.13 & 0.590%
& 3.02 & 0.036 & 1.21 & 0.006 & 34.36 & 0.584 & 59.18 & 0.436 \\
Age-conditioned Neural ODE
& 3.08 & 0.066 & 1.22 & 0.010 & 34.49 & 0.452 & 59.30 & 0.379%
& \underline{2.98} & 0.035 & 1.20 & 0.006 & \underline{34.32} & 0.770 & \underline{59.15} & 0.577%
& 3.02 & 0.037 & 1.21 & 0.005 & 34.38 & 0.552 & 59.21 & 0.426 \\
MoE Neural ODE
& 3.08 & 0.071 & 1.22 & 0.011 & 34.54 & 0.648 & 59.33 & 0.521%
& 2.98 & 0.034 & \underline{1.20} & 0.008 & 34.31 & 0.826 & 59.14 & 0.601%
& \underline{3.02} & 0.038 & \underline{1.21} & 0.006 & \underline{34.39} & 0.634 & \underline{59.21} & 0.473 \\
Stage Neural ODE
& 3.08 & 0.062 & 1.22 & 0.009 & 34.51 & 0.411 & 59.45 & 0.382%
& 3.01 & 0.035 & 1.21 & 0.007 & 33.63 & 0.938 & 58.69 & 0.640%
& 3.04 & 0.037 & 1.21 & 0.006 & 33.98 & 0.681 & 58.99 & 0.486 \\
\midrule
DevelopmentODE (ours)
& \textbf{2.93} & 0.062 & \textbf{1.21} & 0.010 & \textbf{37.51} & 0.404 & \textbf{61.56} & 0.344%
& \textbf{2.94} & 0.034 & \textbf{1.20} & 0.005 & \textbf{35.26} & 0.784 & \textbf{59.82} & 0.577%
& \textbf{2.94} & 0.032 & \textbf{1.20} & 0.005 & \textbf{36.16} & 0.517 & \textbf{60.51} & 0.386 \\
\bottomrule
\end{tabular}%
}
\end{table*}

\subsection{Full Independent Validation Results}
\label{Indepent}
We additionally evaluate all compared models on the external BCP cohort to
examine whether the performance patterns observed in the main experiments also
hold in a distinct cohort. Table~\ref{tab:external_comparison} reports the full
comparison across the evaluated baselines.


DevelopmentODE achieves the lowest MSE and highest PCC among the evaluated
methods on this cohort. These results provide complementary evidence for the
effectiveness of the proposed framework under an additional cohort setting.

\begin{table*}[t!]
\centering
\caption{
Comparison on the external BCP cohort.
MSE and PCC are multiplied by $100$ for readability.
The best results are in \textbf{bold}, and the second-best
are \underline{underlined}.
}
\label{tab:external_comparison}

\small
\setlength{\tabcolsep}{4.5pt}
\renewcommand{\arraystretch}{1.08}

\resizebox{\textwidth}{!}{%
\begin{tabular}{lccccccccccc}
\toprule
\textbf{Metric}
& \textbf{RNN}
& \textbf{GRU}
& \textbf{LSTM}
& \textbf{Transformer}
& \textbf{Direct MLP}
& \textbf{Time-cond. MLP}
& \textbf{Generic NODE}
& \textbf{Age-cond. NODE}
& \textbf{MoE NODE}
& \textbf{Stage NODE}
& \textbf{DevelopmentODE} \\
\midrule

MSE $\downarrow$
& 4.7677
& 4.8221
& 4.8894
& 4.7032
& 4.6728
& \underline{4.6256}
& 5.0102
& 5.0071
& 5.0075
& 5.0134
& \textbf{4.0467} \\

PCC $\uparrow$
& 46.3742
& 46.1490
& 46.0146
& 47.0882
& 48.3735
& \underline{48.6872}
& 44.3411
& 44.4017
& 44.3976
& 44.5096
& \textbf{53.6233} \\

\bottomrule
\end{tabular}%
}
\end{table*}

\subsection{Sensitivity to a Learnable Population-Aligned Component}
\label{app:learnable_parallel_stretch}

The main DevelopmentODE formulation uses the population developmental canal
$\mu_\eta(t)$ to define a local developmental direction and models
subject-specific deviation dynamics primarily in its complementary space.
This provides an explicit separation between population-level progression and
the higher-capacity deviation field. We additionally examine a nested
relaxation of this geometric constraint to assess whether retaining a
population-aligned component changes predictive performance.

Using the population tangent $u(t)$ defined in Eq.~(3), we write the
population-aligned and complementary components of a vector $v$ as
\begin{equation}
    P_{\parallel}(v;t)
    =
    \langle v, u(t) \rangle u(t),
    \qquad
    P_{\perp}(v;t)
    =
    v - P_{\parallel}(v;t).
\end{equation}
We then define a one-parameter relaxation
\begin{equation}
    P_{\alpha}(v;t)
    =
    P_{\perp}(v;t)
    +
    \alpha P_{\parallel}(v;t),
    \label{eq:learnable_parallel_projection}
\end{equation}
where
\begin{equation}
    \alpha = \tanh(a),
\end{equation}
and $a$ is a single learnable scalar initialized at zero.
The original complementary-space formulation is recovered at $\alpha=0$,
whereas nonzero $\alpha$ retains a controlled component along the population
developmental direction. As $\alpha$ approaches one, both the complementary
and population-aligned components are retained.

For this variant, $P_{\alpha}$ replaces the complementary-space operation
applied to both the subject deviation and the corresponding deviation-field
outputs. Thus, a nonzero $\alpha$ allows the deviation field to retain
population-aligned state information and to express a population-aligned
velocity component, while preserving the same population-referenced
decomposition used by DevelopmentODE.
We evaluate the relaxed formulation under the same all-forward and adjacent
transition protocols as the fixed complementary-space model. Within each protocol,
the two variants use identical subject-disjoint five-fold splits,
fold-specific PCA transformations, initialization, optimization settings,
validation-based checkpoint selection, and evaluation pairs. The only
difference is whether the population-aligned component is fixed to zero or
controlled by the learned coefficient $\alpha$.

\begin{table}[t]
\centering
\scriptsize
\setlength{\tabcolsep}{2.5pt}
\renewcommand{\arraystretch}{0.95}
\caption{
Sensitivity to a learnable population-aligned component.
Results are mean $\pm$ SD over five subject-disjoint folds.
MSE, $R^2$, and PCC are $\times100$; MAE is $\times10$.
}
\label{tab:learnable_parallel_stretch}
\resizebox{\linewidth}{!}{%
\begin{tabular}{llccccc}
\toprule
Pairs & Model & $\alpha$
& MSE $\downarrow$
& $R^2$ $\uparrow$
& MAE $\downarrow$
& PCC $\uparrow$ \\
\midrule

\multirow{2}{*}{All}
& Fixed comp.
& $0$
& $2.9403 \pm 0.0326$
& $36.128 \pm 0.509$
& $1.2009 \pm 0.0046$
& $60.495 \pm 0.383$ \\

& Learnable aligned
& $0.153 \pm 0.088$
& $2.9390 \pm 0.0321$
& $36.158 \pm 0.527$
& $1.2005 \pm 0.0052$
& $60.515 \pm 0.396$ \\

\midrule

\multirow{2}{*}{Adjacent}
& Fixed comp.
& $0$
& $2.9451 \pm 0.0262$
& $36.534 \pm 0.321$
& $1.2033 \pm 0.0041$
& $60.773 \pm 0.256$ \\

& Learnable aligned
& $0.198 \pm 0.025$
& $2.9428 \pm 0.0266$
& $36.586 \pm 0.307$
& $1.2032 \pm 0.0042$
& $60.808 \pm 0.242$ \\

\bottomrule
\end{tabular}%
}
\end{table}

As shown in Table~\ref{tab:learnable_parallel_stretch}, the learned
coefficient remains modest in both settings, with
$\alpha=0.153\pm0.088$ for all-forward pairs and
$\alpha=0.198\pm0.025$ for adjacent pairs. The relaxed model therefore retains
a limited population-aligned component rather than remaining exactly at the
$\alpha=0$ solution.
The resulting changes in aggregate forecasting performance are small.
Relative to the fixed complementary-space formulation, the mean MSE decreases
by approximately $0.043\%$ in the all-forward setting and $0.078\%$ in the
adjacent setting. The corresponding changes in the other metrics are similarly
small, and are substantially smaller than the variation observed across
subject-disjoint folds.
These results indicate that allowing a limited population-aligned component
does not materially change the forecasting behavior of DevelopmentODE under
the present evaluation settings. At the same time, the learned nonzero
coefficients show that the model can make use of a modest relaxation when it
is available. We therefore retain the complementary-space formulation in the
main model because it provides a simpler population--individual decomposition
while achieving essentially the same predictive performance.

\section{Limitations}

(1) DevelopmentODE is designed as a structured predictive framework for learning continuous developmental dynamics from sparse longitudinal observations. The developmental canal, complementary geometry, and age-dependent modulation provide a principled way to translate hypotheses about shared developmental progression, individual variability, and developmental non-stationarity into explicit inductive structure for longitudinal forecasting. Beyond improving prediction, this formulation offers a structured representation for examining how population-level progression and individual variation may be organized across development. Our analyses show that the learned components exhibit systematic developmental patterns, while establishing their causal or mechanistic correspondence to underlying neurodevelopmental processes remains an important direction for future work. (2) Our empirical evaluation focuses on resting-state functional connectivity during early childhood. Longitudinal neuroimaging over multi-year developmental periods is inherently sparsely sampled, and datasets with dense repeated measurements spanning a full decade remain limited. Within this setting, our primary cohort covers the first decade of life, and we additionally evaluate DevelopmentODE on the independently collected BCP cohort, where the overall performance advantage is preserved. BCP covers a shorter developmental interval, so this experiment provides complementary evidence of cross-cohort robustness within early development. As larger longitudinal resources become available, evaluation across additional developmental periods, imaging modalities, and clinical populations will enable a broader characterization of the applicability and generality of the proposed framework. (3) Several modeling choices are intentionally kept simple to isolate the contribution of the proposed dynamical structure and to accommodate the limited longitudinal supervision available in current neuroimaging datasets. In particular, directly learning both the developmental representation and stage organization from sparse and irregular longitudinal observations would introduce substantially greater flexibility while providing relatively limited supervision for reliably identifying these components. We therefore use a fold-specific PCA representation and a small number of ordered developmental stages with fixed transition centers. As larger and denser longitudinal datasets become available, future work could jointly learn the developmental representation and stage organization with stronger longitudinal supervision.

\end{document}